%% file: ms.tex
\pdfoutput=1
\documentclass{article}
\usepackage{preprint}
\usepackage[twoside,letterpaper,includeheadfoot,%
        layoutsize={8.125in,10.875in},%
                layouthoffset=0.1875in,%
                layoutvoffset=0.0625in,%
                left=1.25in,%
                right=1.25in,%
                top=45pt,%
                bottom=45pt,%
                headheight=0pt,% No Header
                headsep=10pt,%
                footskip=25pt]{geometry}
\usepackage[numbers, compress]{natbib}
\usepackage{chngcntr}
\usepackage{graphicx,amsmath,amsfonts,amssymb}
\usepackage{booktabs}
\usepackage[utf8]{inputenc} % allow utf-8 input
\usepackage[T1]{fontenc}    % use 8-bit T1 fonts
\usepackage{hyperref}       % hyperlinks
\usepackage{microtype}      % microtypography
\usepackage{graphicx}
\usepackage[font=footnotesize]{caption}

\usepackage[utf8]{inputenc} % allow utf-8 input
\usepackage[T1]{fontenc}    % use 8-bit T1 fonts
\usepackage{hyperref}       % hyperlinks
\usepackage{url}            % simple URL typesetting
\usepackage{booktabs}       % professional-quality tables
\usepackage{amsfonts}       % blackboard math symbols
\usepackage{nicefrac}       % compact symbols for 1/2, etc.
\usepackage{microtype}      % microtypography

\usepackage{amsmath} % Math utilities.
\usepackage{comment}
\usepackage[capitalise,noabbrev]{cleveref} % Smart References
\usepackage{xspace}
\usepackage{xcolor}
\usepackage{color} % Colored text
\usepackage{subcaption} % Subfigures.
\usepackage{url}		
\usepackage{graphicx}
\usepackage{color}    % Colored text.
\usepackage{svg}		% Small sized table / figure captions.
\usepackage{wrapfig}
\usepackage[skip=2pt]{caption} % Smaller figure captions
\usepackage{listings} % latex in verbatim

\input{macros.tex}

\title{CURI: A Benchmark for Productive Concept Learning Under Uncertainty}

\author{%
\large
  \textbf{Ramakrishna Vedantam}$^1$,
  \textbf{Arthur Szlam}$^1$,
  \textbf{Maximilian Nickel}$^1$,\\
  \textbf{Ari Morcos}$^1$,
  and
  \textbf{Brenden Lake}$^{1,2}$\\
  \normalsize $^1$Facebook AI Research; \normalsize $^2$New York University
}

\date{}

\begin{document}
\maketitle

\input{sections/abstract}

\section{Introduction}
\label{sec:intro}
\input{sections/intro_post_neurips}

\section{Related Work}
\label{sec:related}
\input{sections/related_new.tex}

\section{Productive Concept Learning (\pcl) Dataset}
\label{sec:datasets}
\input{sections/datasets.tex}

\section{Metrics and Baselines}
\label{sec:metrics}
\input{sections/baselines.tex}

\section{Experimental Results}
\label{sec:experiments}
\input{sections/experiments.tex}

\section{Conclusion}
\label{sec:conclusion}
\input{sections/conclusion.tex}

\section*{Acknowledgments}
\label{sec:ack}
\input{sections/ack.tex}

\bibliographystyle{chicago}
\bibliography{abstraction,library_clean}

\clearpage

\section*{Appendix}

\appendix
\input{sections/appendix.tex}

\end{document}

%% file: macros.tex
\newenvironment{packed_itemize}{
\begin{list}{\labelitemi}{\leftmargin=2em}
\vspace{-6pt}
 \setlength{\itemsep}{0pt}
 \setlength{\parskip}{0pt}
 \setlength{\parsep}{0pt}
}{\end{list}}

\newcommand{\rviclr}[1]{\textcolor{black}{#1}}

\newcommand{\avgpool}{avg-pool\xspace}
\newcommand{\relnet}{relation-net\xspace}
\newcommand{\cat}{concat\xspace}
\newcommand{\trnsf}{transformer\xspace}

\newcommand{\versus}{\textit{v.s.}\xspace}

\newcommand{\pcl}{CURI\xspace}
\newcommand{\cba}{CBA\xspace}
\newcommand{\cgap}{\texttt{comp gap}\xspace}
\newcommand{\bindcolor}{Binding (color)\xspace}
\newcommand{\bindshape}{Binding (shape)\xspace}
\newcommand{\intrinsic}{Intrinsic\xspace}
\newcommand{\extrinsic}{Extrinsic\xspace}
\newcommand{\complexity}{Complexity\xspace}
\newcommand{\bool}{Boolean\xspace}
\newcommand{\comp}{Concept IID\xspace}
\newcommand{\counting}{Counting\xspace}
\newcommand{\iid}{Instance IID\xspace}
\newcommand{\map}{mAP\xspace}

\mathchardef\mhyphen="2D

\newcommand{\myvec}[1]{\mathbf{#1}}

\newcommand{\vc}{\myvec{c}}

\newcommand{\vu}{\myvec{u}}

\newcommand{\vx}{\myvec{x}}

\newcommand{\vI}{\myvec{I}}

\newcommand{\mymathcal}[1]{\mathcal{#1}}
\newcommand{\calG}{\mymathcal{G}}
\newcommand{\calH}{\mymathcal{H}}

\newcommand{\calL}{\mymathcal{L}}

\newcommand{\calT}{\mymathcal{T}}
\newcommand{\calU}{\mymathcal{U}}

\newcommand{\calY}{\mymathcal{Y}}

\newcommand{\etc}{\emph{etc.}\xspace}
\newcommand{\ie}{\emph{i.e.}\xspace}
\newcommand{\eg}{\emph{e.g.}\xspace}

\newcommand{\union}{\cup}
\newcommand{\intersect}{\cap}

\newcommand{\eat}[1]{}
\newcommand{\remark}[2]{} %{\textcolor{blue}{#1}: \textcolor{blue}{#2}}
\newcommand{\replace}[2]{} %{\sout{#1} \textbf{#2}}

\newcommand{\stefan}[1]{}%\todo[linecolor=blue,backgroundcolor=blue!10,bordercolor=blue]{SML: #1}}

%% file: sections/abstract.tex
\begin{abstract}
Humans can learn and reason
under substantial uncertainty in a space of
infinitely many concepts, including structured relational concepts
(``a scene with objects that have the same color'') and ad-hoc categories defined through goals (``objects that could fall on one's head''). In contrast, standard classification benchmarks: 1) consider only a fixed set of category labels, 2) do not evaluate compositional concept learning and 3) do not explicitly capture
a notion of reasoning under uncertainty. We introduce a new few-shot, meta-learning benchmark, Compositional Reasoning Under Uncertainty (\pcl)
to bridge this gap. \pcl evaluates different aspects of productive and systematic generalization, including abstract understandings of disentangling,
productive generalization, learning boolean operations, variable binding, etc. Importantly, it also defines
a model-independent ``compositionality gap'' to evaluate difficulty of generalizing out-of-distribution along each of these axes.
%Further, PCL can be rendered using three different modalities (images, schemas and sound), emphasizing abstract reasoning over particular modalities.
Extensive evaluations across a range of modeling choices spanning different modalities
(image, schemas, and sounds), splits,
privileged auxiliary concept information,
and choices of negatives reveal substantial scope for modeling advances on the proposed task.
%including models trained with and without auxiliary concept descriptions. We hope that our work will spur more research on human-like concept learning in unbounded, productive concept spaces.
All code and datasets will be available online.

\end{abstract}

%% file: sections/intro_post_neurips.tex
% People can reason in more productive and flexible ways than today's AI systems, understanding completely novel concepts and expressions \citep{Lake2017-pp,Ho2019-gt}.
% \ari{We still don't have a definition for productive anywhere here. I really think we have to define it in some way. Alternative suggestion: "People can reason flexibly, using a limited set of experiences to produce and reason about a large set of novel expressions and concepts, yet today's AI systems lack this ability."}
Human concept learning is more flexible than today's AI systems. Human conceptual
knowledge is \emph{productive}: people can understand and generate novel concepts via compositions of existing concepts (``an apartment dog'') \citep{Murphy2002}, unlike standard machine classifiers that are limited to a fixed set of classes
(``dog'', ``cat'', etc.).  Further, humans can induce goal-based, ``ad hoc'' categories such as
``things to take from one's apartment in a fire'' (children, dogs, keepsakes, etc.)~\citep{Barsalou1983-ud}. Thus, unlike AI systems, humans reason seamlessly in large, essentially ``unbounded'' concept spaces.

\begin{figure}
    \centering
    \includegraphics[width=\textwidth]{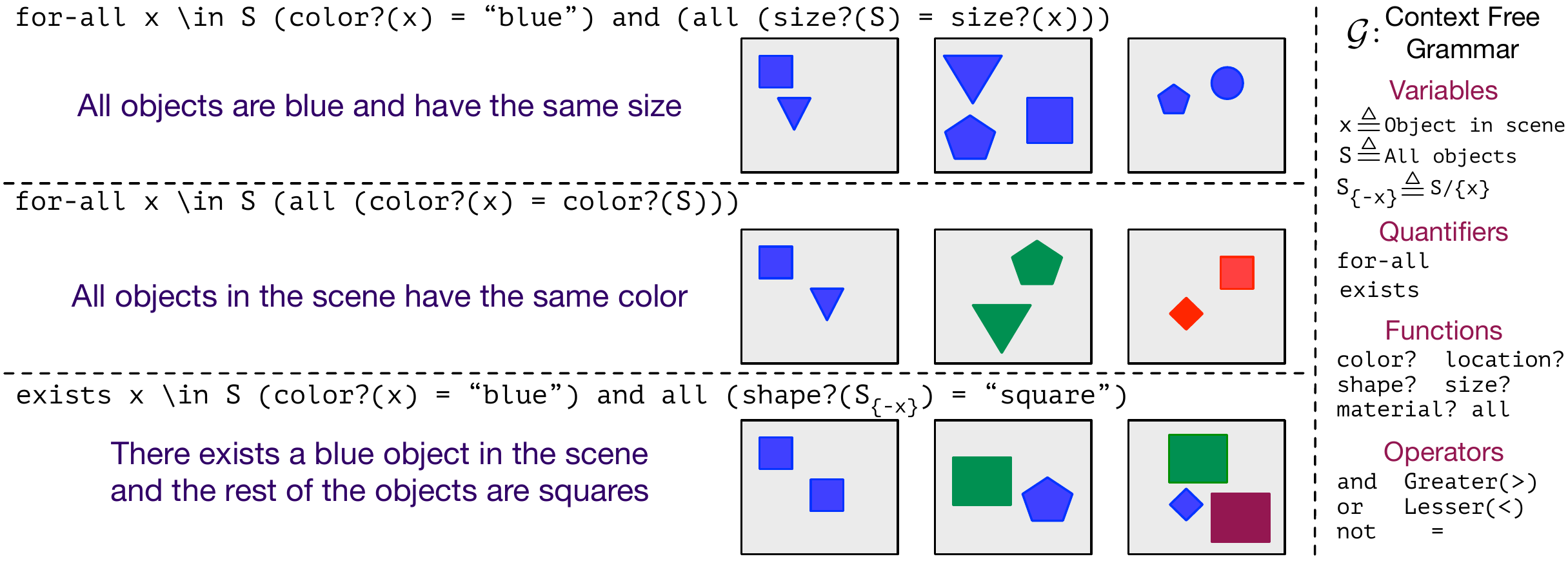}
    \caption{\noindent\textbf{Concept Space}. Three example concepts (rows)  along with schematic positive examples. Actual scenes are rendered  in multiple ways including the CLEVR renderer \citep{Johnson2016_clevr} (see \cref{fig:task}).
    \textbf{Right:} The grammar of variables, quantifiers, functions and operators to induce compositional concepts.}
    \vspace{-10pt}
    \label{fig:teaser}
\end{figure}

Beyond unboundedness, a natural challenge in such concept spaces
is \emph{uncertainty} -- the right concept to be
inferred is uncertain, as a plethora of candidate concepts could
explain observations. For \eg in ~\cref{fig:teaser} (top, image panel),
the ``right'' concept could be that ``All objects are blue and have
the same size'', but it could also be ``There are less than four
objects in the scene'', or ``All objects have the same color''.
Humans gracefully handle such uncertainty and underdetermination \citep{Tenenbaum2001,Xu2007,Goodman2008a,Piantadosi2016-bx}.
Popular compositional reasoning
benchmarks such as CLEVR~\citep{Johnson2016} for
visual question answering and
Ravens Progressive Matrices
~\citep{santoro_neurips2017} for
deductive, analogical reasoning are
compositionally rich and challenging
in nature, but do not tackle ambiguity and
underdetermination.

%\begin{figure}[tbp]
%\begin{subfigure}{\textwidth}

%\caption{Elements of the Compositional Reasoning Under Uncertainty (\pcl) task.}
%\vspace{-50pt}
%\end{figure}

We address this gap in 
the literature, and propose
the Compositional Reasoning Under Uncertainty
(\pcl) benchmark to study how modern machine learning systems can
learn concepts spanning a large,
productively defined space (\cref{fig:teaser}). In pursuit of this goal, we instantiate
a meta learning task where a model must acquire a compositional
concept from finite samples. A signature of productivity in human thought
is our ability to handle novel combinations of known,
atomic components. Thus, in \pcl we instantiate different
systematic train-test splits
to analyze different forms of generalization in concept learning, involving novel combinations of intrinsic properties (\eg color, shape) with boolean operators, counting, extrinsic object properties (\eg object location), and a novel test of variable binding in context of compositional learning. 

While related
systematic splits have been proposed in prior work in context
of other tasks such as question answering and analogical
reasoning
~\citep{Barrett2018-qr,Hill2019-as,Agrawal2017-qw,Johnson2016_clevr,imagination,Higgins2017-wa,Bakhtin2019-sy,LakeBaroni2018,Ruis2020}, ours is the first benchmark which tests different qualitative
aspects of reasoning about productive concepts under uncertainty.

\begin{wrapfigure}{r}{0.5\textwidth}
    \begin{center}
    \vspace{-10pt}
    \includegraphics[width=0.48\textwidth]{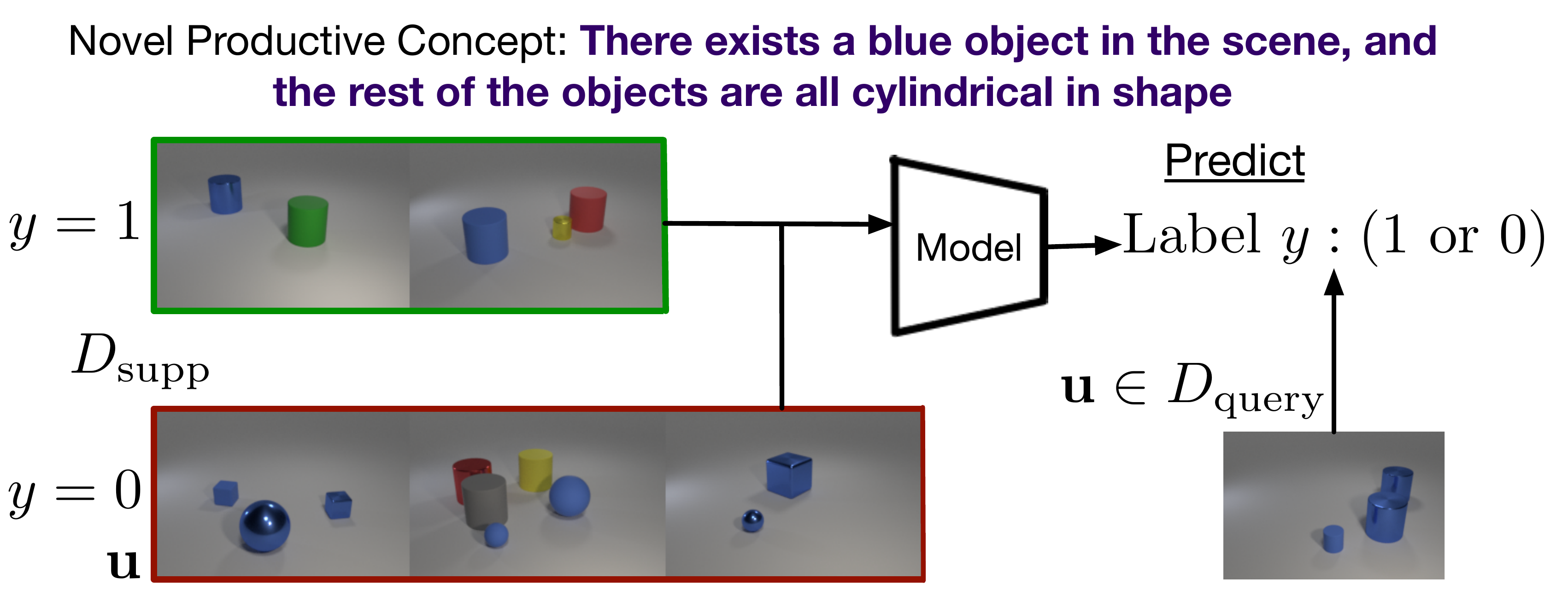}
    \caption{\textbf{\pcl Task}. Given as input a support
    set $D_{\text{supp}}$, with positive and negative examples
    corresponding to concept, the model has to infer the
    concept and produce
    accurate predictions on novel images (right).}
    \label{fig:task}
    \vspace{-15pt}
    \end{center}
\end{wrapfigure}

\noindent\textbf{Compositional Reasoning Under Uncertainty (\pcl) Task.} Concretely,
the
\pcl task
tests few-shot learning of relational concepts in a large compositional conceptual
space, with design inspiration
from studies in cognitive modeling using
a language of thought (LOT) approach~\citep{Fodor1975,Piantadosi_thesis,Kemp2005}. \pcl includes scene-based concepts such as ``All objects have the same color'' and ``There exists a blue object while the rest are triangles'' (\cref{fig:teaser})
but unlike CLEVR \citep{Johnson2016_clevr}
there are too few examples to deduce answers with certainty. Our benchmark is defined through a series of meta-learning episodes (see example in \cref{fig:task}): given positive and negative examples of a new concept $D_{\text{supp}}$ (known as the ``support set''), the goal of an episode is to classify new examples $D_{\text{query}}$ (the ``query set'').
As in few-shot classification \citep{FeiFeiFergus2006}, meta-learning \citep{Vinyals2016-ri}, and other open-set tasks \citep{Lampert2014-qo}, models are evaluated on novel classes outside the (meta-)training set. Unlike previous work~\citep{meta_dataset,LakeOmniglotProgress} that focuses on atomic concepts, our benchmarks concerns more structured, relational concepts built compositionally from a set of atomic concepts, and involves
reasoning under uncertainty -- an ideal learner must marginalize over many hypotheses when making predictions \citep{Gelman2004,Xu2007,Piantadosi2016-bx}.

We also vary the modality in which scenes are presented---rendering them as images, symbolic schemas, and sounds--- \rviclr{enabling future research on modality-specific representational choices
for compositional reasoning under uncertainty.}
%further emphasizing abstract reasoning over the particulars of the input representation. 
Finally, we vary the concepts learned by the model during meta-training and meta-testing to test different aspects
of systematic generalization.

\noindent\textbf{Compositionality Gap.} In addition to defining systematic splits, we also characterize (for
the first time, in our knowledge), the difficulty
of generalization entailed by each split by introducing the notion
of a model-independent ``compositionality gap''. Concretely, the compositionality gap is the difference in test performance between an ideal Bayesian learner with access to the full hypothesis space, and a Bayesian learner with access to only a (potentially large) list of the hypotheses examined during meta-training. A large gap indicates that any learner must extrapolate compositionally from the training hypotheses to solve the task; additionally, models can be compared to ideal learners that either do or do not engage in such extrapolation. We anticipate that this tool will be more broadly useful for analyzing other benchmarks with compositional splits.

\noindent\textbf{Models.} We evaluate models
around various dimensions which concern
the difficulty of learning productive concepts under uncertainty, including:
1) the modality in which the input is rendered (image, schemas, sounds), 
2) method used for reasoning across objects in a scene \rviclr{(\trnsf, relation-network, global average pooling, concatenation)},
3) whether or not training provides ground-truth symbolic descriptions of concepts,
and 4) how negative examples are sampled. 
%Our models are instantiated
%in the formalism of~\citet{Snell2017-rl} --
%a simple yet state-of-the-art meta-learning model
%~\citep{meta_dataset}. 
%operating over different architectures for reasoning about objects in scenes. We also evaluate models trained to predict classification labels and symbolic (LOT) descriptions at training time, and those that are only trained to predict classification labels.
Overall, our evaluations suggest that there is
substantial room for improvement in compositional
reasoning under uncertainty, w.r.t the compositionality gap, representing
a novel challenge for compositional learning.

\noindent\textbf{Summary of contributions:}
1) We introduce the Compositional Reasoning Under Uncertainty (\pcl) benchmark for evaluating compositional, relational learning under uncertainty from observational data; 
2) We introduce a `compositionality gap' metric for measuring the difficulty of systematic generalization from train to test;
3) We provide various baseline models for \rviclr{benchmarking progess}.

%% file: sections/related_new.tex
\noindent\textbf{Compositional Learning.} Related work has examined systematic generalization in pattern completion using Raven's matrices (PGM) ~\citep{santoro_neurips2017,Hill2019-as} and visual question answering with CLEVR~\citep{Johnson2016_clevr,Bahdanau2019}. \pcl's use of the CLEVR renderer further invites particular comparison with that benchmark. Compared to these more deductive reasoning tests, \pcl examines few-shot concept learning under substantial inherent uncertainty. Unlike puzzle solving or question answering, an ideal inductive learner on \pcl cannot know the right rule with certainty. In essence, unlike CLEVR
the ``question'' to be answered
is not given to the model as input, but
must be inferred -- making the task
more challenging. 
While PGMs do involve such an inference, once the constraints of a puzzle are identified, it does not: 1) have any uncertainty in the reasoning (which is crucial) and 2) involve any “concept” learning – where a concept applies to multiple images – as much as it involves “instance” matching to complete a sequence.
In contrast, a successful \pcl model behaves as if marginalizing over many hypotheses consistent with the observations e.g., \citep{Tenenbaum2001,Xu2007,Piantadosi2016-bx}, an ability which is rarely studied directly in deep learning models (although see \citep{Grant2019a}).
%Beyond the aspect of uncertainty,
%raven's
%Additionally, while both the PGM~\citep{Barrett2018-qr,Hill2019-as} and CLEVR~\citep{Johnson2016_clevr,Bahdanau2019} datasets initially proved challenging for modern models, subsequent work has achieved impressive performance on the out-of-distribution splits present in both challenges~\citep{Hill2019-as, santoro_neurips2017}, suggesting that more dificult tasks such as \pcl are needed.

Recently,~\citet{Keysers2019-yu} proposed
a method to create ``difficult'' systematic
splits based on the principle that they should share atoms but have
maximally different compositions. This
is complementary to our splits, which
provide interpretable notions of what
each split tests such as disentangling,
complexity, variable binding \etc Moreover, our variable binding split is
predicated on having different atoms
between train and test, and thus cannot
be recovered by their methodology.

%In other related work, \citet{Andreas2017-yp} studied few-shot learning with concepts defined by language-like structured descriptions. Models were trained to treat language as a latent variable (or bottleneck), inducing novel classifiers from examples via the space of linguistic expressions. Our goals differ in evaluating systematic generalization in productive concept learning in models that do not have access to the latent language, although we compare with language-augmented models as well.

\noindent\textbf{Language of Thought (LOT).} Our choice of compositional concepts was most closely inspired by \citep{Piantadosi2016-bx} along with other studies of human concept learning in the Language of Thought (LOT) framework~\citep{Fodor1975,Goodman2008a,Kemp2009-sn,Piantadosi2012,Goodman2014,Overlan2017,LakeFractals2019}. In typical LOT studies of human learning, the conceptual space $\calH$ is defined through a probabilistic context-free grammar $\calG$, which specifies a set of conceptual primitives and their rules of combination. 
%Not only does $\calG$ generate the stimuli, it is often used for cognitive modeling; learning is operationalized as a Bayesian search through the LOT for the best descriptions of the examples. Our goals differ here; 
Here, we use a LOT-inspired grammar $\calG$ to generate an unbounded set concepts $\calH$, while evaluating machine learning models trained without access to the underlying LOT.

%% file: sections/datasets.tex
\begin{figure}
    \centering
    \includegraphics[width=\textwidth]{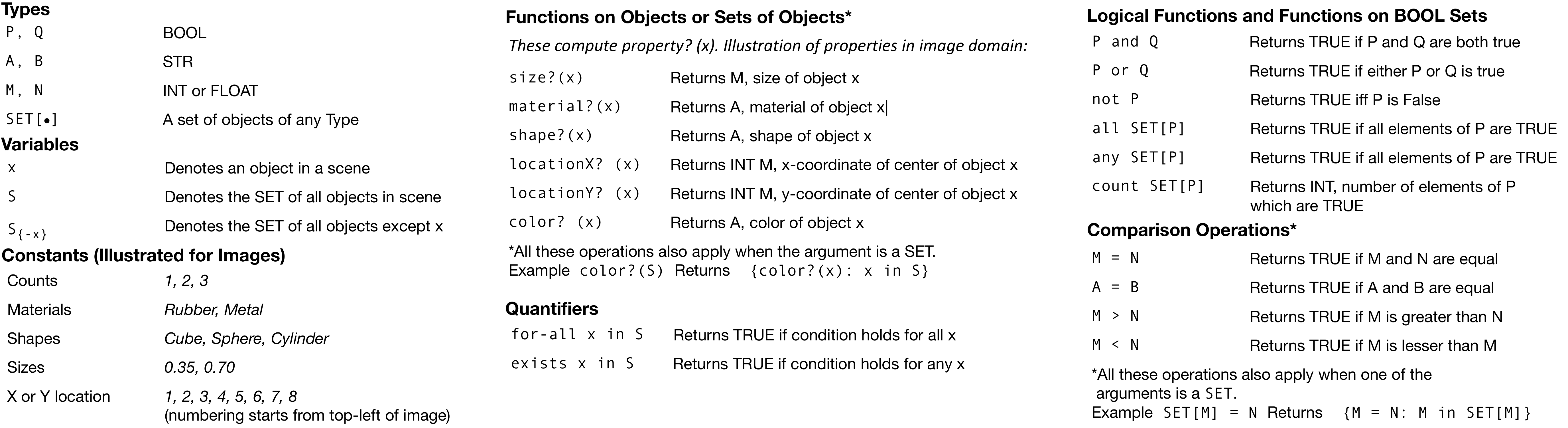}
    \caption{\textbf{Language of thought.}
    All valid (type-consistent) compositions of functions are potential complex
    concepts in our dataset. Note that the functions are illustrated
    for the case of images and schemas. Location, size, shape \etc
    correspond to different properties for sounds.} 
    \label{fig:lot_figure}
\end{figure}

\noindent\textbf{Concept space.}
The compositional concepts in \pcl were inspired by the empirical and cognitive modeling work of~\citet{Piantadosi2016-bx}.
The space of concepts (LOT) is defined by a context free grammar ($\calG$). \cref{fig:lot_figure} shows the LOT and specifies how primitives and functions compose to produce a large unbounded concept space. The LOT has three variables: $\vx$, representing an object in a scene, $S=\{\vx\}_{i=1}^{N}$ representing the set of all objects in the scene, and $S_{-\vx}=S/\{\vx\}$, representing the set of all objects in the scene \textit{except} $\vx$. Each concept describes a rule composed of object and scene properties, logical operators, and/or comparison operators, and can be evaluated on a given scene $S$ to determine whether the scene satisfies the rule. 

Object and scene properties are defined by functions which can be applied to objects or scenes: for example, $\texttt{size?}(\vx)$ yields the size of an object $\vx$, while $\texttt{size?}(S)$ returns a set with the sizes of all the objects ($\{\texttt{size?}(\vx): \vx \in S\}$).
Comparison and logical operators can be used to compare and relate various properties of objects in scenes.
In contrast to~\citet{Piantadosi2016-bx}, we include a $\texttt{count}$ operator, which determines how many times a condition is satisfied by a set, which allows us to check how well deep learning models are able to count~\citep{Chattopadhyay2016-bg,Johnson2016_clevr,Agrawal2017-qw}.
Finally, quantifiers such as $\texttt{exists}$ and $\texttt{for-all}$ enrich the LOT by specifying the number of objects which must satisfy a given condition.
% This rich grammar allows us to sample myriad compositional concepts for testing abstraction capabilities of machine learning systems (see~\cref{fig:teaser} for examples)\footnote{Additional details are provided in Appendix \ari{Add appendix section}}.

Consider the following example concept (\cref{fig:teaser} bottom): ``There exists a blue object in the scene and the rest of the objects are squares.'' To access the color of a given object, we use \texttt{color?($\vx$)} and to access the shape of a given object, we use \texttt{shape?($\vx$)}. To determine whether an object matches a specific property, we can combine this with equality: \texttt{shape?($\vx$) = ``square''}. Finally, we can use \texttt{exists} to specify that at least one object must be blue, $S_{-\vx}$ to specify all the objects except for that blue object, and \texttt{all} to specify that all the objects in $S_{-\vx}$ must be squares. Putting it all together: \texttt{exists $\vx \in S$ (color?($\vx$) = ``blue'') and all (shape?($S_{-\vx}$) = ``square'')}.

% Given a sampled concept, $h \sim \calG$, namely
% \texttt{$\lambda$ S. for-all x $\in$ S (all (color? (x) = color? (S)))}, one can read it
% as follows. The first part \texttt{$\lambda$ S.} denotes that we are writing
% a function of a scene $S$. Next, we iterate over all \texttt{x}
% $\in$ \texttt{S} and evaluate the inner expression \texttt{all (color? (x) = color? (S))}.
% This means that for a given choice of \texttt{x}, we evaluate if its color matches the color
% of ``all'' the objects in the scene \texttt{S}, and then want this condition to hold true
% for all objects \texttt{x} in the scene. In English, this is the concept ``All objects
% in the scene have the same color.''. \brenden{I suggest starting this paragraph with English expression rather than ending with it. Then explain how you get the interpreted.}

% Overall, we sample and filter a large set of concepts to obtain a set of 14,929
% concepts $\calH$ for training and evaluating machine learning models (more details
% can be found in the appendix). 
% Given this base set of concepts,
% we next explain different partitions of the concept space
% that yeild the generalization
% regimes we explore in the paper.

\noindent\textbf{Structured Generalization Splits.}
% Productivity in humans is related to
% our ability to understand and learn
% concepts and expressions that involve novel combinations
% of known primitives~\citep{Fodor1975},
% such as functions, properties and logical operations. In this paper, our goal is to benchmark the
% capability of machines to perform such
% productive concept learning.
A signature of productivity is the ability to handle novel combinations of known components~\citep{Fodor1975,Fodor1988}.
Thus, in \pcl, we consider splits that require generalizing to novel combinations of known elements from our LOT (\cref{fig:lot_figure}), including \rviclr{combinations} of constants, variables, and functions. We achieve this by creating disjoint splits of concepts $\calH_{train}$ 
and $\calH_{test}$ for training and evaluating models. By varying the held out elements and their combinations, we obtain splits that evaluate different axes of 
generalization.
In practice, we use our grammar $\calG$ to
sample and filter a large set of concepts (see~\cref{subsec:sampling} for more details), which yields a set of 14,929 concepts $\calH$ for training and evaluation. We next describe how each
split divides $\calH$ into $\calH_{train}$ and $\calH_{test}$, \rviclr{to test productive, out of distribution generalization:}

\begin{packed_itemize}
    \item \textbf{Instance IID}: Evaluates generalization to novel episodes from the same concept set. This is the standard setup in machine learning~\citep{murphy2013machine}, in which $\calH_{train}$ = $\calH_{test}$. This is the only split where train and test concepts overlap.
    \item \textbf{Concept IID}: Evaluates generalization to novel concepts based on an arbitrary random split of the concepts into $\calH_{train}$ and $\calH_{test}$.\footnote{
    While some strings $h$ might be different in surface form,
    they may yeild the same results when applied to images. In this
    split we account for such synonomy, and ensure that no two
    concepts which are synonyms are in different splits. See~\cref{sec:synonomy}
    for more details.}
    % \ari{I think Brenden's point from before about this still holds now that we've removed statement from previous paragraph. Pointing this out only for Concept IID makes it seem like $\calH_{train} \intersect \calH_{test} = \emptyset$ is only true for Concept IID. We should move the statement back to the previous paragraph I think. It's helpful that Instance IID specifies that it's the only non-disjoint split, but I think is still confusing}
    %
    \item \textbf{Counting}: Evaluates the ability to learn a new concept $h$ with novel property-count combinations, e.g, the training concepts never filter for exactly `3 squares'.
    \item \textbf{Extrinsic properties}: Evaluates the ability to learn a new concept $h$, with novel combinations of extrinsic (e.g. location) and intrinsic (e.g. color) object properties.
    
    \item \textbf{Intrinsic properties}: Evaluates the ability to learn a new concept $h$ with novel combinations of intrinsic properties, e.g., the training concepts never reference both `red' and `rubber'.
    % For example, concepts in the training set might include `green' and `metallic`, or `red' and `rubber' in them, while the test set might involve learning of a concept
    % with `red' and `metallic' in the concept string $h$.
    %
    \item \textbf{Boolean operations}: Evaluates the ability to learn concepts which require application of a familiar boolean operation to
    a property to which the operation has never been applied previously.   %
    \item \textbf{Complexity split}: Evaluates generalization from simple concepts (those which have less than or equal to 10 symbols) to more complex concepts (longer than 10 symbols). This
    is indicative of the productivity~\citep{Fodor1975} exhibited by models, in generalizing from simpler concepts to more complex concepts.
    \item \textbf{Variable binding}: Evaluates learning of entirely novel intrinsic properties, e.g. the training concepts involve only ``red'', ``blue'', and ``green'' but test concepts involve ``yellow'' (although `yellow' objects can still appear in training scenes).
    This is indicative of inferential coherence~\citep{Fodor1975} in models, in generalizing rules of inference to novel atoms.
\end{packed_itemize}

A model that infers the underlying LOT during meta-training would be expected to perform well on any such systematic split. By comparing the performance of current models to to such ideal learners, this benchmark will allow us to evaluate progress on the systematic out-of-distribution generalization capabilities of our current models.~\cref{sec:structured_gen_splits} provides
more details on the strucutred splits.

% \brenden{Not sure what this is current paragraph is aiming to day. Can we simply say that any of these splits would be solvable by a model that extrapolates the underlying LOT from the 14K training episodes? And we test an ideal learner that does that, and one that doesn't, in order to see how models compare (comp. gap).} While the systematic splits divide concepts along all these different axes
% for generalization, the task itself is not impossible in principle in terms
% of the information available to models. For the \textbf{color} split, for
% example, even though one is never asked to learn a concept involving \texttt{red}
% during training, one still observes the concept red in the input modality provided
% to the model in $D_{supp}$. Thus, across all the systematic generalization
% settings investigated in our work, one still `sees' the held out concept/
% primitives. The notion of generalization is in having to learn or `abstract'
% a novel compound concept from (potentially) similar observations. We believe
% that more broadly, this is a promising setting to make progress on in tasks
% related to compositionality and abstraction.
\label{subsec:episodes}
\noindent\textbf{From Concepts to Meta-learning Episodes.} A single episode comprises a support set ($D_{\text{supp}}$) and a query set ($D_{\text{query}}$), each of which is generated from a given concept, $h$. Formally,
a support or query set $D$ has input data $\vu$ and corresponding
label $y$, \ie $D = \{\{y_i\}_{i=1}^N, \{\vu_i\}_{i=1}^{N}\}$. Each support and query set contains 5 positive and 20 negative examples --- negative examples are oversampled since the space of negatives is generally much larger than that for positives. The set of positive examples are sampled uniformly from a categorical distribution over all positives. However, we consider two types of negatives: 1) easy negatives, in which the negatives are also sampled at random, and 2) hard negatives, in which negatives are generated from a closely related concept which also evaluates true on the positive examples in $D_{supp}$, such that these negatives are maximally confusing. Altogether, for each split, \rviclr{our} train, validation, and test sets \rviclr{contain} 500000, 5000, and 20000 episodes, respectively.

\noindent\textbf{Compositionality Gap.}
A key aspect of our benchmark is to define the difficulty \rviclr{in learning}
that arises from the compositional structure of the concept space. Most of the splits above are structured in a way such that
$\calH_{\text{test}} \intersect \calH_{\text{train}} = \emptyset$ -- forcing a learner
to use the compositional structure of the concept space to generalize to
$\calH_{\text{test}}$. We conceptualize the difficulty of this task through the
notion of its \emph{compositionality gap}. Intuitively, the compositionality gap
captures the difference between the generalization performance of an ideal
compositional learner (\emph{strong oracle}) compared to an ideal
non-compositional learner that is unable to extrapolate outside the training
concepts (\emph{weak oracle}).

Formally, let $\Omega \in \{\text{strong}, \text{weak}\}$ denote an oracle over
a concept space $\calH_{\Omega}$. The posterior predictive distribution of an
oracle for query scene $\vu$ and query label $y\in\{0,1\}$ is then given as:
$p_{\Omega}(y| \vu, D_{\text{supp}}) = \sum_{h \in \calH_{\Omega}} p_{\Omega}(y| h, \vu) p_{\Omega}(h| D_{\text{supp}})$, 
where $p_{\Omega}(h| D_{\text{supp}}) \propto p_{\Omega}(h) \text{ } p(\{y_i\}_{i=1}^N|h; \{\vu_i\}_{i=1}^{N})$ and $p_{\Omega}(h)$ denote the
posterior and prior, respectively. Given a metric of interest $M$ (e.g., mean
average precision or accuracy), the compositionality gap of a learning task
is then simply defined as the difference in performance of the strong and
weak oracle when evaluating on concepts from $\calH_{\text{test}}$, i.e., $M(p_{\text{strong}}) - M(p_{\text{weak}})$.

Using this notion of compositionality gap, we can then define ideal learners,
i.e., the strong and weak oracle,
simply via their priors. In particular, let $w(h)$ denote a weight on importance of each hypothesis\footnote{
Set to log-linear in the prefix serialization
length of hypothesis, inspired by the observation that longer
hypotheses are more difficult for humans~\citep{Feldman2000-on}. See~\cref{sec:weight_prior} for more details.} and let $\vI$
denote the indicator function. We then define the prior of an oracle as
$p_{\Omega}(h) = \sum_{h'\in\calH_{\Omega}} w(h') \vI[h'=h],
  % p_{so}(y| \vu, D_{supp}) & = \sum_{h \in \calH} p(y| h, \vu) p_{so}(h| D_{supp}).
$. The difference between strong and weak oracle lies in which
concepts can be accessed in these priors.

In this formalism, the strong oracle has access to the union of train and test concepts; that is  $\calH_{\text{strong}} = \calH_{\text{train}} \cup \calH_{\text{test}}$. The weak oracle, on the other hand only assumes access to $ \calH_{\text{weak}} = \calH_{\text{train}}$,
%to a prior $p_{\calH_{train}}(h) = \sum_{h'\in\calH_{train}} w(h') \vI[h'=h]$,
which means it is unable to consider any hypothesis outside what has been seen
in training and assigning it zero probability mass. Given a support set
$D_{\text{supp}}$ this difference in priors leads then to different inferences
on posteriors and allows us to quantify how compositionally novel a learning
task is relative to these ideal learners.

%% file: sections/baselines.tex
During meta-test, given $D_{\text{supp}}$ models are 
evaluated on their ability to learn novel concepts. We use two metrics for quantifying
this:
1) \textbf{Accuracy:} 
% Given a support set $D_{supp}$ provided to the model at validation or test
% time, we would like to evaluate the degree to which it understands the compositional concept. As explained
% in~\cref{sec:intro}, we take a view of evaluating this through the quality of predictions made by the model.
evaluates the accuracy of model predictions across the query set $D_{query}$, as is standard practice in meta-learning~\citep{LakeOmniglotProgress,Snell2017-rl}. Since there are more negative than positive labels, we report class balanced accuracy for better interpretability, averaging accuracies for the positive and negative query examples; and 2) \textbf{mean Average Precision (mAP):} evaluates models on a much larger number of test scenes $\calT$ for each episode (comprising 44,787 scenes, 3 per each concept in $\calH$).
This resolves an issue that with a small query set, a strong model could achieve perfect accuracy without grasping the concept. Since episodes typically have many more negative than positive examples, Average Precision sweeps over different thresholds of a model's score and reports the average of the precision values at different recall rates, e.g., ~\citep{pascal_voc}. mAP is then the mean across all of the meta-test episodes. 

\begin{figure}
    \centering
    \includegraphics[width=\textwidth]{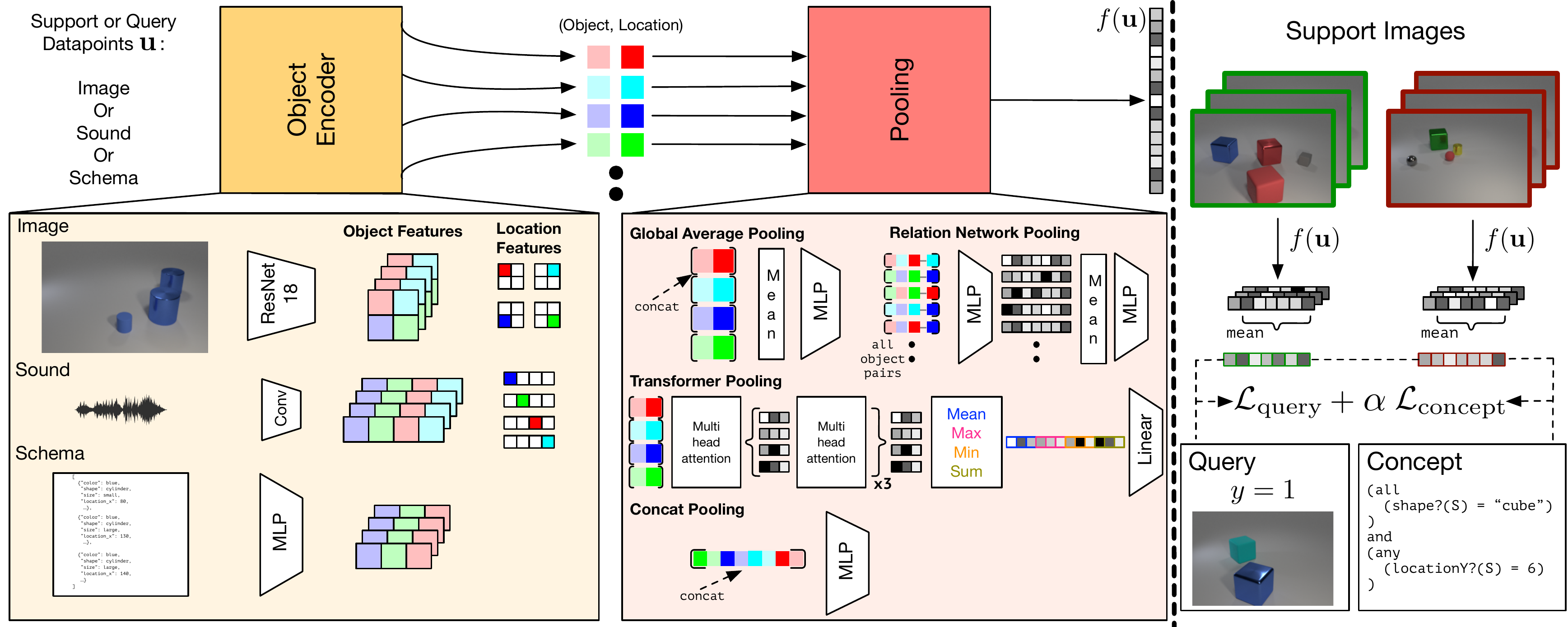}
    \caption{\noindent\textbf{Baseline models (left).} Different choices
    for the encoder $f(\vu)$ parameterization explored in the paper.
    We consider three modalities each of which is processed with a
    modality specific encoder, followed by four kinds of pooling architecture
    which take as input objects and their corresponding locations
    to provide an encoding for the datapoint. \textbf{Training
     (right).} The model is trained by processing the support
     images $D_{\text{supp}}$ with positive (green) and negative (red)
     images, using $f(\vx)$ to compute $\calL_{\text{query}}$ which computes
     generalization error on queries and $\calL_{\text{concept}}$ which
     learns to
     decode the true concept as an auxiliary task. Losses
     are weighted by $\alpha \ge 0$.}
     \label{fig:model_figure}
\end{figure}

\subsection{Training Loss}\label{subsec:losses}
%\brenden{This setup below is completely standard, and I suggest we cut this whole section. After we describe the model architecture, we can just say we do maximum likelihood training.}
Denote by $\vu \in \mathbb{R}^M$ the input to the model, which can be either in the form of
image, sound or schema.  We work in a binary classification setting with
labels $y$ live in the space $\calY \in \{0, 1\}$.  Then, given a support set $D_{\text{supp}} = \{\vu_i, y_i\}_{i=1}^{T}$ and a query set $D_{\text{query}}=
\{\vu_i, y_i\}_{i=1}^{T}$, sampled in accordance with a productive concept $h$, our training objective
for a single training instance
can be written as  $ \calL_{\text{query}}  + \alpha \calL_{\text{concept}}$.  Here
 $\calL_{\text{query}} = \sum_{\vu, y \in D_{\text{query}}}\log p(Y=y|\vu, D_{\text{supp}})$
is a standard maximum likelihood meta-learning loss ~\citep{Ravi2016-vg,
Snell2017-rl,Finn2017-bw}, and $\calL_{\text{concept}} = \log p(H=h| D_{\text{supp}})$
is an optional regularizer designed to encourage retaining information about the hypothesis of interest from the support set.  

\subsection{Baseline Model Architectures}
Our baseline models (shown in \cref{fig:model_figure}) parameterize the probability in the $\calL_{\text{query}}$ term \rviclr{above} using prototypical networks~\citep{Snell2017-rl}.
%, which is a particularly strong yet simple meta learning approach. 
The prototypical network consists of 
an embedding function $f = f_{\theta}$ 
%: \mathbb{R}^M \rightarrow \mathbb{R}^N$ where  for datapoints $\vu \in D_{supp}$
and uses it to compute prototypes $\vc_p$ and $\vc_n$ for positive and negative examples %(\cref{fig:model_pooling (right)})
by averaging $f(\vu)$ for positive and negative examples in the support set respectively.  In equations,
%Let us denote the prototype for positive examples as $\vc_{1}$, and negatives as $\vc_{0}$, and $d(\vx, \vy) = ||\vx - \vy||^2$, the squared euclidean distance between $\vx, \vy \in \mathbb{R}^N$.
%Then, 
given a query datapoint $\vu'$, we compute  %~\citet{Snell2017-rl} 
%compute $z' = f_{\theta}(\vu')$ and parameterizes the conditional distribution of the label $y$ as~\citep{Snell2017-rl}:

\vspace{-4mm}
\begin{equation}
p(Y=y| \vu; D_{\text{supp}}) = \frac{exp( ||f(\vu')-\vc_p||^2}{ exp( ||f(\vu')-\vc_p||^2) +exp( ||f(\vu')-\vc_n||^2 }
\end{equation}

\rviclr{In this formalism, the models we study in this paper span different choices for $f$.} Roughly, in each modality, we start with an encoder that converts the raw input into a set of vectors, and then a pooling operation that converts that set of vectors into a single vector.  In the case of images and sound (input as spectrograms), the encoder is a ResNet-18; and the set of vectors is a subsampling of spatial locations; and for schemas we vectorize components with a lookup table and combine them into a set via feed-forward networks.  In the case of images and sounds, the output of the encoder is enriched with position vectors. For the pooling operation, we study global averaging, concatenation, relation networks~\citep{santoro_neurips2017} \rviclr{and transformers~\citep{Vaswani_transformer} equipped
with different pooling operations (max, mean, sum, min) for reasoning inspired by~\citet{Wang2019-tb} (\cref{fig:model_figure} middle panel, also see~\cref{sec:detailedarch} for more details).}

For the probability in \rviclr{$\calL_{\text{concept}}$}, \rviclr{we represent the concept as a sequence by prefix serialization and then use} an LSTM~\citep{LSTM} to parameterize
$p(h| D_{\text{supp}}) = \Pi_{s=1}^S p(h_s| h_{1\cdots t-1}; D_{supp})$.
%to parameterize the density $p(h| D_{supp}) = \Pi_{s=1}^S p(h_s| h_{1\cdots t-1}; D_{supp})$,
At each step of the LSTM we concatenate $[\vc_p, \vc_n]$ to the input.

%% file: sections/experiments.tex
\begin{figure}[t]
    \centering
    \includegraphics[width=\textwidth]{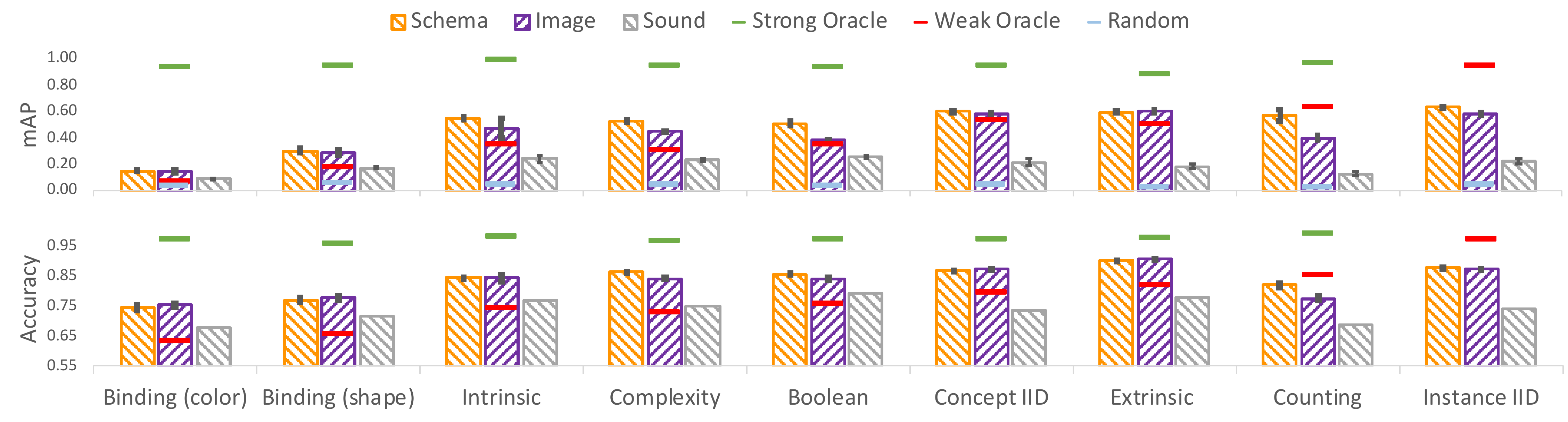}
    \caption{\textbf{Compositionality Gap}.
    Different splits (x-axis)
    plotted w.r.t performance of the strong oracle (green line) and
    weak oracle (red line) on the \map (top)
    and Accuracy (bottom) evaluated
    on respective test splits (using hard negatives in support and query sets).
    Difference between the two is the compositionality gap (\cgap).
    \textbf{Yellow:} shows the (best) relation-net model
    on schema inputs,
    \textbf{purple:} shows the model on image inputs, and \textbf{gray:} shows the model on sound inputs.
    Error bars are $\texttt{std}$ across 3 independent model runs.
    \label{fig:results_rel_net}}
\end{figure}

We first discuss the compositionality gap induced by the different generalization
splits and then delve into the impact of modeling choices
on performance on the generalization splits.
All models are trained
for 1 million steps,
and are run with
with 3 independent training runs to report standard
deviations.
We sweep over 3 modalities (image, schema, sound),
\rviclr{4} pooling schemes (\avgpool, \cat, \relnet, \rviclr{\trnsf}),
2 choices of negatives (hard negatives, random negatives)
and choice of language ($\alpha=0.0, 1.0$). Unless
mentioned otherwise in the main paper we focus on results
with hard negatives and $\alpha=0.0$. When instantiated
for a given modality, we note that the encoders $f(\vu)$ (\cref{fig:model_figure}) all have
a similar number of parameters. The appendix contains details of the exact
hyperparameters (\cref{sec:reproducibility}), and more comprehensive
results for each split (\cref{subsec:full_results}).

\subsection{Dataset Design and Compositionality}

\textbf{How compositional are the structured splits?}.
Our main results are shown in \cref{fig:results_rel_net}.
Using our model-independent measure of the compositionality gap (Section \ref{sec:datasets}), different splits present varying challenges for generalizing from train to test. The most difficult splits, with the largest compositionality gaps, are the \bindcolor and \bindshape, which is reasonable since they require learning concepts with entirely new property-values. In contrast, the easiest split with the smallest compositionality gaps is the \iid split
since it does not require compositionality. Finally, while the \map
metric exposes a larger value of \cgap, the ordering of splits in terms
of \cgap is same for both metrics -- suggesting similar coarse-grained notions of compositionality.

Results for the best overall architecture, a relation network (\relnet), is shown in \cref{fig:results_rel_net}. Network performance on the easiest data format (schema; yellow bars) is generally better than the weak oracle, but substantially worse than the strong oracle. Counting is a particularly challenging split where the models underperform even the weak oracle. Broadly, this suggests that the models
capture some notion of compositionality -- especially
for images and schemas -- relative to a weak
oracle that rigidly considers only training hypotheses, but there is substantial room to improve (especially with respect to the
more stringent \map metric). These results demonstrate that \pcl provides a challenging yet tractable setting for evaluating the compositional capabilities of models.

Finally, we found
that the performance on
the \iid split is not equal to the weak (and strong) oracle---which are both equal in this case---indicating that the best model does not make ideal posterior predictions even when compositionality is not an issue. Ideal predictions in this case would require the network to behave as if marginalizing over the training hypotheses, as the strong oracle does. A similar plot to~\cref{fig:results_rel_net} can be found in~\cref{subsec:easynegatives}
for random negatives. 

\textbf{Influence of Negatives}. Previous work~\citep{Hill2019-as}
has shown that the choice of random \versus hard negatives for training and evaluation impacts compositional generalization
substantially in the case of a particular set of analogical reasoning models. However, we argue that such decisions on dataset
design can be made more objectively if one can evaluate the model-independent \cgap. 
In our context, we find that the \cgap with \map when using random negatives
decreases on
average by 5.5 $\pm$ 1.4\% compared to when we use hard negatives.
This indicates that it is not only the choice of $\calH_{train}$ and $\calH_{test}$,
which are identical for a given compositional split (say \counting), but
also the choice of the negatives which ``makes'' the task compositionally
novel. More
generally, this indicates that the \cgap has utility as a more general
diagnostic tool for making
principled design decisions in compositional learning settings \textit{without} the confound of specific model decisions.

\subsection{Differences Between Models}
\textbf{Best Models}. In general, the best performing model is the \relnet applied to schema inputs, outperforming other combinations of models
and input modalities on the \bool, \comp, \complexity, and \iid splits
on both the \map as well as accuracy metrics (\cref{fig:results_rel_net}); although as mentioned above, none of the models are close to the strong oracle.
\rviclr{It is closely followed by the \trnsf model on schema inputs, which 
performs the best on \bindcolor, \bindshape, and \intrinsic splits (\cref{subsec:full_results}).}
Utilizing schema inputs proves easier for abstraction except
for the \extrinsic setting, where the task requires generalization to novel
locations for objects in images, which is well supported by the inductive bias of
the CNN encoder (\cref{fig:model_figure}).
In this case, the \rviclr{image-\trnsf} get\rviclr{s} an \map of
\rviclr{$62.1 \pm 0.7$\%}, \rviclr{compared to the next best schema-\trnsf model
at $60.9 \pm 0.7$.}
\rviclr{Further, r}elational learning proves more crucial in the schema case \rviclr{
than for images, with all image models (regardless of pooling) performing better
than $59.4 \pm 1.3\%$ \map (achieved for image-avg-pool) while schema-avg-pool models get only get to $53.4 \pm 1.5 \%$.}

\textbf{When to use a \rviclr{transformer}?} 
Transformer models appear to outperform relation networks in splits
concerning disentangling. For instance, for the \intrinsic
split with schema-relation-net is at \rviclr{$55.1 \pm 0.8$\% \versus $57.9 \pm 0.6$\%
for schema-\trnsf. Similarly, for the \extrinsic split the image-\trnsf is at $62.1 \pm
0.7$\% compared to the image-relation-net at $60.8 \pm 1.1$\%. We
hypothesize that this is because the iterative
message passing via. attention in transformers improves object representations for disentangling compared to relation networks that lack
such a mechanism.}

\textbf{What is the relative difficulty of abstraction from different
modalities?}
One of the key contributions of our work is in providing multiple modalities
(image, schema, sound) for productive concept learning. We next
characterize the difficulty of abstraction based on modality for the various
generalization settings.
In the \intrinsic
setting, we find that the schema models, which have access to a ``perfect''
disentangled representation significantly outperform image models---a schema-avg-pool model gets an \map of $52.7\pm3.1$\% while
an image-avg-pool model gets to $34.4\pm0.0$\% \map.

Similarly, for the
\counting split where the total number of objects are exactly specified
in the schema (\cref{fig:model_figure}), schemas
are substantially better than images. For example, schema-relation-nets
get to $56.25\pm5.32$\% \map while
image-avg-pool is at $48.4\pm1.2$\% \map. Interestingly,
the next best model---image-relation-net---is substantially worse,
at $39.45\pm1.6$\%. \rviclr{Curiously, for while \trnsf models perform
well at disentangling, they seem to be quite poor for \counting, with
image-\trnsf models getting to only $32.4 \pm 1.4$\% \map, suggesting
a potential weakness for transformers.}
Overall, there appears to be an intimate link between the generalization
setting and the input modality\rviclr{, suggesting avenues where representation
learning could be improved for a given modality (e.g. images), relative to the kind of
reasoning one is interested in (e.g. counting).}

%Both variable binding splits were challenging for all modalities (\cref{fig:results_rel_net}), yet the baseline models were still superior to the weak oracle.  Note that the image
%distributions are the same across training and test, meaning that the ``held
%out'' color (say red) has been ``seen'' in images at training time, even though it has not been used in the training hypotheses.

%Interestingly, for both variable binding splits \bindcolor and \bindshape we find that the task is equally challenging using either image or schema inputs (\cref{fig:results_rel_net}). At a first glance it might appear surprising that despite obvious difficulty of variable binding, models perform better than the weak oracle (\cref{fig:results_rel_net}). We hypothesize this is because the image distributions are the same across training and test, meaning that the ``held out'' color (say red) has been ``seen'' in images at training time, but never been explicitly ``learnt''. Hence, the difficulty is not in bottom up processing of a red pixels, but in top-down ``re-cognition'' of the red and associating it to the abstract notion of `color'.

\textbf{When does language help?} 
%We next test if by providing formal concept definitions during training time, in addition to the query set, one can imbue a baseline model with the language of thought.
On average, training models with explicit concept supervision using the
concept loss (\cref{subsec:losses}) improves performance by $2.8 \pm 0.6$\% \map (SEM error). This is a small boost relative to the gap between the 
original model
and the strong oracle, suggesting that this simple auxiliary loss is not sufficient to internalize the LOT in a neural network. Overall, image models benefit more from language than schema models which natively utilize symbols (\cref{subsec:languageuse}).
%\textbf{Impact of negatives on the learning of the model}
%Finally, we evaluate models trained with random negatives at test time
%on the more compositional hard negatives set\todo{We find that..}.

%% file: sections/conclusion.tex
We introduced the compositional reasoning under uncertainty (\pcl) benchmark for evaluating few-shot concept learning in a large compositional space, capturing the kinds of producitivity, unboundness and \rviclr{underdetermination} that characterize human conceptual reasoning. 
We instantiate a series of meta-learning tasks, and evaluate numerous baseline models on various aspects of compositional reasoning under uncertainty, including inferential coherence, boolean operation learning, counting, disentangling,
\textit{etc.} Further, we introduce
the notion of a compositionality gap to quantify
the difficultly of each generalization type, and to estimate the degree of compositionality in current deep learning models. 
%We also provide
%a comprehensive set of models and baselines for future work to compare to. 
We hope our contributions of dataset, compositionality gaps, evaluation metrics and baseline models help spur progress in the important research direction of productive concept learning.

%% file: sections/ack.tex
We would like to thank Laurens Van Der Maaten, Rob Fergus, Larry Zitnick, Edward Grefenstette, Devi Parikh
and numerous other colleagues at Facebook AI
Research for their feedback and discussions in helping shape this project. Specifically,
we thank Larry Zitnick and Edward Grefenstette for comments on this draft. Finally, we would like to thank the developers of Hydra and PyTorch for providing amazing frameworks for running large scale deep learning experiments.

%% file: sections/appendix.tex
\section{Example episodes from the dataset}
We show examples from the \comp split test set comprising the ground truth
productive concept (top), along with
the support and query sets for meta learning (rendered as images),
the alternate hypotheses which are consistent with
the support set -- that is, other hypotheses which could also have generated
the positive and negative examples in the support set -- and the concepts
based on which
we pick the hard negatives \cref{fig:qual_1,fig:qual_2,fig:qual_3,fig:qual_4,fig:qual_5,fig:qual_6}.

\input{sections/appendix_sections/qualitative_examples.tex}

\section{Additional Dataset Details}
We first provide more details of the concept space $\calG$, then explain
how we obtain $\calH$, the space of concepts for training and evaluation,
provide more details of the structured splits, and finally explain the
weight $w(h)$ based on which we sample concepts.

\subsection{More details of the grammar}
We provide below the full grammar used to sample concepts, where
$A \rightarrow B| C$ means that $A$ can expand to either $B$ or $C$
under the rules defined by the grammar. We always start expanding
at the \texttt{START} token and then follow the rules of the grammar
until we hit a terminal node (which does not have any expansions
defined). As and where possible, we followed the insights 
from~\citet{Piantadosi2016-bx} in choosing the sampling probabilities
for various completions based on how well humans seem to be able to
learn the corresponding primitive. For example, we sample utterances
with disjunctions (\texttt{or}) less frequently since they are known
to be difficult for humans to learn. Based on~\citet{Kemp2009-sn},
we chose to represent location as a discrete entity, such that
relative, and categorical notions of \texttt{left} or \texttt{right}
simply become comparisons in the location space ($\text{location? } \vx
> \text{location? } S_{-\vx}$), unlike the CLEVR
dataset \citep{Johnson2016_clevr} which defines categorical relational
objects.

Here is the full grammar $\calG$ used for sampling the concepts (as
explained in the main paper, $S_{-\vx} = S /\{\vx\}$).
Note that the grammar always generates strings in postfix notation and thus
the operands in each expansion occur before the operation:

\input{sections/appendix_sections/full_grammar.tex}

\subsection{Sampling}\label{subsec:sampling}
We sample 2000000 initial hypotheses from the CFG $\calG$,
and impose a maximum depth in the recursion tree of 6 when sampling. That is, no node 
has a depth larger than 6 in the recursion through which we generate concepts from
the grammar $\calG$. We then reject and filter the hypotheses to obtain a
set of ``interesting'' hypotheses
$\calH$ used in the main paper explained in more detail below:

\noindent\textbf{Rejection Sampling:} We reject the following string combinations after sampling
from the grammar $\calG$:
\begin{itemize}
    \item All programs which contain "$\lambda S.$ for-all x" and "$S_{-\vx}$" in the same program. This is
    asking that for all objects in a scene, a certain property is satisfied by everything other
    than the object, which is the same as saying, for all objects in the scene. 
    \item All programs where we compare the properties of the same object to itself, e.g. 
    \texttt{color? (x) == color? (x)}, where color? can be any function applied to the object. 
    \item All programs where we have the following string: \texttt{exists(color?(S) == color?(x))} or
    \texttt{for-all(color?(S) == color?(x))} where color? can be any function applied to the object. 
    \item All programs which evaluate to true on schemas more than 10\% of the time and less than
    10 times. The former condition ensures that we work with concepts which are in some sense
    interesting and surprising (as opposed to concepts which are always trivially true),
    and the second condition ensures that we have unique schmeas or datapoints to place in
    the support and query sets, which both have 5 positive images each.
\end{itemize}

We provide examples of concepts which get rejected for being true too often below:
\input{sections/appendix_sections/removed_concepts.tex}

See \cref{sec:structured_gen_splits} for more details on the structured generalization
splits which yeild train concepts $\calH_{\text{train}}$ and test concepts $\calH_{\text{test}}$.

\subsection{Concept prior weight $w(h)$}\label{sec:weight_prior}
We next explain the form of the prior weight $w(h)$ that we use for defining the prior
over the concepts provided to the models (both oracles as well as deep learning models).
Given $l(h)$, the number of tokens in the postfix serialization of the concept $h$,
the unnormalized weight $\tilde{w}(h)$ is log-linear in the length, and is defined as follows:
\begin{equation}
    \tilde{w}(h) \propto \exp{- 0.2 \cdot l(h)}
\end{equation}

Given a split $\Omega \in \{train, test\}$, the final, normalized weight is given
as:
\begin{equation}
    w(h) = \frac{\tilde{w}(h)}{\sum_{\calH_{\Omega}} \tilde{w}(h)}
\end{equation}

As explained in the main paper, the final prior for a hypothesis given a split $\Omega$
is $p(h) = \sum_{h' \in \calH_{\Omega}} w(h) \vI[h=h']$.

Our choice of the log-linear weight is inspired by the observation in cognitive science
that longer boolean concepts are harder for people to learn~\citep{Feldman2000-on}.

Here are some examples of hypotheses with a high weight (computed on $\Omega = \text{train} \union \text{test}$):
\input{sections/appendix_sections/high_prob_prior.tex}

\subsection{Execution on Images.}\label{sec:execution}
In order to create the perceptual inputs in the dataset $\calU$,
we sample images using the renderer for CLEVR 
from~\citet{Johnson2016_clevr}, changing the range of objects to $[2, 5]$,
to reduce clutter and enable easier learning of predicates like
\texttt{any} and \texttt{all} for models.
\footnote{Since the chances of a constraint being true for all obejcts
reduce exponentially as the number of objects increases.}
The CLEVR renderer produces scenes $\vu$ with
pixels as well as an associated schema file $\vu^s$ detailing the properties
of all the objects sampled in the scene, including their location,
shape, size, material, and rotation. Based on this, we convert our
sampled concepts into
postfix notation and execute them on the schemas using
an operator stack. Concretely, execution of the concept $h \in \mathcal{H}$
on $\vu^s$ yields a boolean true or false value $\{0, 1\}$.
We execute each such hypothesis on a set of 990K images, yielding
scores of how often a hypothesis is true for an image. We threshold
this score to retain the subset of hypotheses which are true
for no more than 10\% of the images and are true at least for 10 images,
to pick a subset of ``interesting'' hypotheses $\mathcal{H}'$ for training
models.

\begin{figure}
    \centering
    \includegraphics[width=\textwidth]{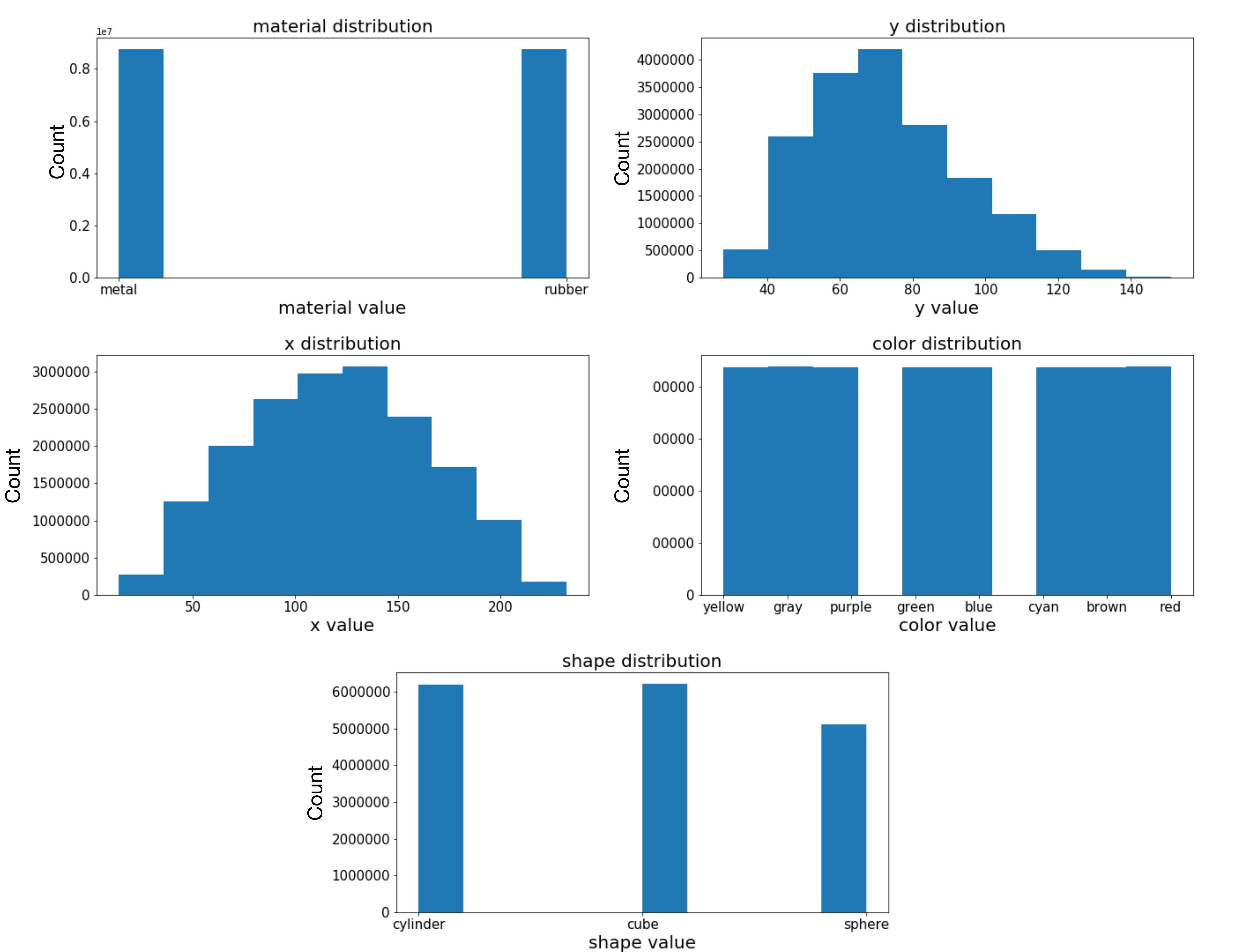}
    \caption{Histogram of properties found in inputs $\vu$ in the dataset. We notice that
    the properties are all largely uniform with bias in x and y-coordinates towards the center
    of the image.\label{fig:property_distribution}}
\end{figure}

\textbf{Bias.}
The image dataset here sampled itself has a bias in terms of the location coordinates (in the pixel
space). The CLEVR dataset generation process samples objects in the 3d (top-down x, y) coordinate
space uniformly (from a grid of -3, to +3).
However, since the camera is always looking into the scene from outside, the image formation
geometry implies that in the camera / image coordinates most of the objects appear to be away
from the scene and very few are close to the camera. Thus, in terms of the y-coordinates we observe
in the image coordinates a bias in terms of the distribution not being unifrom. This also makes
sense in general, as even in the real world, objects are not found very close to the camera or
very far away from the camera in general. See \cref{fig:property_distribution} for all
the biases in the observation space $\vu$ computed over 990K sampled images.

\subsection{Audio.}\label{sec:audio}
To build the audio data, we use clips of orchestral instruments playing various pitches downloaded from \url{https://philharmonia.co.uk/resources/sound-samples/}.  We make the following mappings of object properties:
\begin{itemize}
    \item $x$ location $\rightarrow$ temporal location.  larger $x$ bin means the note is played later.  
    \item $y$ location $\rightarrow$ pitch.  All pitches between the instruments are the same (up to octaves).
    \item color $\rightarrow$ instrument
    \begin{itemize}
        \item gray $\rightarrow$ trumpet
        \item red $\rightarrow$ clarinet
        \item blue $\rightarrow$ violin
        \item green $\rightarrow$ flute
        \item brown $\rightarrow$ oboe
        \item purple $\rightarrow$ saxaphone
        \item cyan $\rightarrow$ french-horn
        \item yellow $\rightarrow$ guitar
    \end{itemize}
    \item shape $\rightarrow$ amplitude profile; either getting louder, getting softer, or constant volume
    \item size $\rightarrow$ total volume
    \item material $\rightarrow$ low-pass filtering or no filtering.
\end{itemize}
All binned quantities use the same number of bins as in the image domain.

\subsection{Analysis of synonomy of concepts.}\label{sec:synonomy}
\begin{figure}
    \centering
    \includegraphics[width=\textwidth]{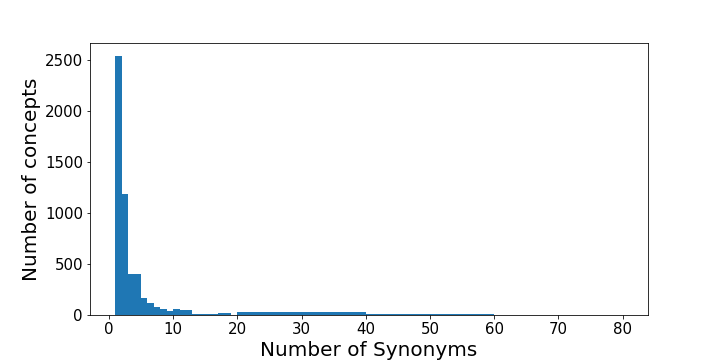}
    \caption{Histogram of number hypotheses with same evaluation signatures.\label{fig:synonym_histogram}}
\end{figure}

We next show an analysis of concepts which have the same evaluation signatures on a large set of 990K
images, and are thus synonymous (in context of the dataset at hand). Note that while some of these
concepts might be truly synonymous to each other (for example, $A>B$ is the same as $B<A$), others
might be synonymous in context of the image distribution we work with. For example, size can never
be greater than 0.7 in our dataset and location can never be greater than 8, and thus asking if location
is greater than 8 or size is greater than 0.7 has the same semantics on our dataset. In \cref{fig:synonym_histogram}
we show each such ``concept'' or ``meaning'', which is a cluster of hypotheses which all evaluate to
the same truth value and plot a histogram of how many hypotheses each cluster tends to have. We notice
that most of the concepts have 1 synonym (i.e. there is only one concept with the particular)
evaluation signature, with a long tail going upto 80 synonyms in a concept.
In the \comp split we ensure that none of the concepts which have the same signature are found
across the train/val/test splits.

\section{Detailed discussion of the structured splits}\label{sec:structured_gen_splits}
We provide more details on how each of the structured splits described in Sec. 3 of the main paper are
created. Assuming access to $\calH$, the space of concepts sampled and filtered from the grammar $\calG$,
we use various heuristics to produce the generalization splits in the paper:
\begin{itemize}
    \item \iid: This split is trivial since $\calH_{\text{train}} = \calH_{\text{test}} = \calH$
    \item \comp: This split divides concepts into train and test by picking concepts at random from $\calH$
      and assigning them to $\calH_{\text{train}}$ or $\calH_{\text{test}}$ while ensuring that no two concepts which
      are synonyms~\cref{sec:synonomy} are found in different splits.
    \item \bool: This split forms cross product of all possible colors and \texttt{\{and, or\}} boolean operators, and
      partitions a subset of such combinations which we want to only occur in test. We use the following tokens for test:
      \begin{verbatim}
        `green', `or' | `purple', `and' | `cyan', `and' |
        `red', `or' | `green', `and'
      \end{verbatim}
      We then create $\calH_{\text{test}}$ to contain all concepts which have any of the combinations above.
      For example, if a concept has both \texttt{green} and \texttt{or} we would place it in $\calH_{\text{test}}$.
      After every feasible candidate is placed in $\calH_{\text{test}}$ based on this heurisitc, the remaining concepts
      in $\calH$ are assigned to $\calH_{\text{train}}$.
    \item \extrinsic: This split forms cross product of all possible colors and locations in the dataset, and partitions
      a subset of such combinations that we only want to occur in test. We use the following tokens for test (only a subset
      shown for illustration):
      \begin{verbatim}
        `7', `gray' | `1', `red' | `3', `purple' |
        `1', `blue' | `8', `cyan' | `5', `yellow' |
        `5', `green' | `3', `yellow' | `7', `purple' |
        `2', `blue' | `3', `cyan'
      \end{verbatim}
      We then create $\calH_{\text{test}}$ to contain all concepts which have any of the combinations above.
      For example, if a concept has both \texttt{gray} and \texttt{7}, and is related to location, that is
      contains \texttt{locationX?} or \texttt{locationY?} keywords, we would place it in $\calH_{\text{test}}$.
      After every feasible candidate is placed in $\calH_{\text{test}}$ based on this heurisitc, the remaining concepts
      in $\calH$ are assigned to $\calH_{\text{train}}$.
    \item \intrinsic: This split forms cross product of all possible colors and materials in the dataset, and partitions
      a subset of such combinations that we only want to occur in test. We use the following tokens for test:
      \begin{verbatim}
       `green', `metal' | `purple', `rubber' | `cyan', `rubber' |
       `red', `metal' | `green', `rubber'
      \end{verbatim}
      We then create $\calH_{\text{test}}$ to contain all concepts which have any of the combinations above.
      For example, if a concept has both \texttt{green} and \texttt{metal}, and is related to material, that is
      contains \texttt{material?} keyword, we would place it in $\calH_{\text{test}}$.
      After every feasible candidate is placed in $\calH_{\text{test}}$ based on this heurisitc, the remaining concepts
      in $\calH$ are assigned to $\calH_{\text{train}}$.
    \item \bindcolor: This split takes all possible colors in the dataset, and partitions
      a subset of colors that we only want to occur in test. We use the following tokens for test:
      \begin{verbatim}
        `purple' | `cyan' | `yellow'
      \end{verbatim}
      We then create $\calH_{\text{test}}$ to contain all concepts which have any of the tokens above.
      For example, if a concept has \texttt{purple}, we would place it in $\calH_{\text{test}}$.
      After every feasible candidate is placed in $\calH_{\text{test}}$ based on this heurisitc, the remaining concepts
      in $\calH$ are assigned to $\calH_{\text{train}}$.
    \item \bindshape: This split takes all possible shapes in the dataset, and partitions
      a subset of shapes that we only want to occur in test. We use the following tokens for test:
      \begin{verbatim}
        `cylinder'
      \end{verbatim}
      We then create $\calH_{\text{test}}$ to contain all concepts which have any of the tokens above.
      For example, if a concept has \texttt{cylinder}, we would place it in $\calH_{\text{test}}$.
      After every feasible candidate is placed in $\calH_{\text{test}}$ based on this heurisitc, the remaining concepts
      in $\calH$ are assigned to $\calH_{\text{train}}$.
    \item \complexity: This split partitions into train and test based on length of the postfix serialization
     of the concept. Specifically, concepts shorter than 10 tokens are placed in $\calH_{\text{train}}$ and longer
     concepts are placed in $\calH_{\text{test}}$.
\end{itemize}

\section{Creating Support and Query Sets}
We next explain how we go from the initial dataset $\mathcal{U}$ -- which
contains a large number of images, schema and sounds -- 
and a concept space $\calH_{\text{train}}$ and $\calH_{\text{test}}$, to a dataset for
meta learning. To create the training/validation/test sets for models, we sample a series of episodes,
each containing a support set and a query set. We illustrate the sampling procedure for a training episode
below:

\noindent{\textbf{Support Set Sampling with Hard Negatives}}
\begin{enumerate}
    \item Pick a concept $h \sim p_{\text{train}}(h)$, with a preference for shorter
    hypotheses being more frequent based on the weights used to define the prior~\cref{sec:weight_prior}
    \item Pick 5 images ($P$), uniformly at random from $\calU$ such that $h(\vu^s)=1$, where the
     concept is evaluated on the schema to determine the label~\cref{sec:execution}
    \item Identify other concepts $h' \in \mathcal{H}$ s.t. $h(u(S)) = 1$ and $h' \neq h$
    \item Pick images such that $h'(\vu^s) = 1$ and $h(\vu^s) = 0$ as negatives ($N$). If no such images
    exist, pick random images from $\calU$ as negatives until we have 20 negatives.
    \item Return $D_{\text{supp}} = P \union N$.
\end{enumerate}

The sampling procedure for the Query set iterates all the steps above (except step 1, where we choose
the concept $h$). Step 3 and 4 outline a procedure for identifying hard negatives for training the
model, by looking at other hypotheses which also explain a chosen set of positives $P$ and using them
to clarify what the concept of interest is.

We give below an analogous procedure for easy 
negatives:

\noindent{\textbf{Support Set Sampling with Easy Negatives}}
\begin{enumerate}
    \item Pick a concept $h \sim p_{\text{train}}(h)$, with a preference for shorter
    hypotheses being more frequent based on the weights used to define the prior~\cref{sec:weight_prior}
    \item Pick 5 images ($P$), uniformly at random from $\calU$ such that $h(\vu^s)=1$, where the
     concept is evaluated on the schema to determine the label~\cref{sec:execution}
    \item Pick 20 random images from $\calU$ as negatives, N.
    \item Return $D_{\text{supp}} = P \union N$.
\end{enumerate}
Similar to hard negatives, the sampling procedure for the Query set iterates
all the steps above (except step 1, where we choose the concept $h$).

\section{Reproducibility and Hyperparameters}\label{sec:reproducibility}
For all the models in Figure. 4 in the main paper, we use the following hyperparameters.
All the modalities are processed into a set of objects $\{o_i\}_{i=1}^{N}$ where 
each $o_i \in \mathbb{R}^{64}$ for image and sound models while for schema $o_i \in \mathbb{R}^{96}$.
Further, the we use a learning rate of \texttt{1e-4} for image models, \texttt{1e-3} for schema
models, and \texttt{5e-5} using the best learning rate for each modality across an initial sweep.
The batch size for image and sound models is 8 episodes per batch, while for schema we use a
batch size of 64. All models use the Adam optimizer. The overall scene representation across
all the modalities is set to 256, that is, $\vu \in \mathbb{R}^{256}$.
All our models are initialized with the method proposed in~\citep{Glorot2010-dm}, which
we found to be crucial for training relation networks well. The initial representation
from the first stage of the encoder (Fig. 4 in main paper) with the objects for images
has a size of \texttt{10x8} \ie there are 80 objects, while for sound representations
have \texttt{38} objects. In the schema case the number of objects is the ground truth
number of objects which is provided as input to the model.

We trained all the models on the training set, and picked the best performing
checkpoints on training -- measured in terms of \map -- to report all the results 
in the main paper. Our image and schema models are trained for 1 million steps
(16 epochs for images, 128 epochs for schemas)
while the sound models are trained for 500K steps, and checkpoints are stored
after every 30K steps.

All our models fit on a GPU with 16GB capacity except the relation network trained with
image inputs, which needs at 32GB GPU. We use the pytorch framework to implement all
our models.

\section{Model Architectures for Pooling}\label{sec:detailedarch}
\rviclr{In this section we detail the exact architectures used for the different
pooling operations we consider in this paper as shown in~\cref{fig:model_figure} center panel.
We first establish some notation. Let $o_i \in \mathbb{R}^K$ be the output
object feature from the modality specific
encoder (\cref{fig:model_figure}, left panel), and let us denote by $O = \{o_i\}_{i=1}^{|O|}$ the set of features for each of the
objects in the scene, which includes optional position information indicating where the object
is present in the scene (\cref{fig:model_figure}). Let $N$ be the
requested dimensionality of the feature space from the pooling operation. Given this, we can describe
the pooling operations used as follows:}

\begin{itemize}
    \item \textbf{avg-pool}: We first average the representations across all the objects
    $\{o_i\}_{i=1}^{|O|}$ and then pass the averaged representation through an MLP with
    \texttt{256 x 512 x 384 x N} units with batch normalization and rectified
    linear unit nonlinearity in the hidden layers.
    \item \textbf{concat}: We first concatenate all the object representations in $O$, followed
    by an MLP with \texttt{256 x 512 x 256 x N} units with batch normalization and rectified
    linear units nonlinearity in the hidden layers.
    \item \textbf{relation-net}: For relation networks, following~\citep{santoro_neurips2017}
    we use relative position encoding that captures the relative positioning of the objects in
    a scene for image and sound modalities, and use the location information already present 
    in the schema modality. Based on this, in the terminology of~\citet{santoro_neurips2017} our
    $g()$ MLP has \texttt{256 x 256 x 256 x 256} hidden units with rectified linear unit non 
    linearity and batch normalization whereas our $f()$ MLP has \texttt{256 x 256 x N} units with
    recitifed linear unit non linearlity and batch normalization in the middle (non-output) layers.
    Different from the original paper, we do not use dropout as we did not observe any overfitting
    in our experiments.
    \item \textbf{transformer}: We use a 2-head multi-head attention layer stacked 4 times,
    with the feedforward network dimenstions set to 512. After forwarding through this module,
    we take the output vectors $o'_i$ for each object processed through these initial layers
    and pool across objects by doing \texttt{max(), mean(), sum(), min()} operations and concatenating
    their outputs, similar to previous work by~\citet{Wang2019-tb}. The final representation then
    does a linear projection of this concatenated vector to $N$, the 
    dimensionality expected from the
    pooling module.
\end{itemize}

\section{Additional Results}
\subsection{Hyperparameter sweeps -- object feature dimensions}
\begin{figure}
    \centering
    \includegraphics[width=\textwidth]{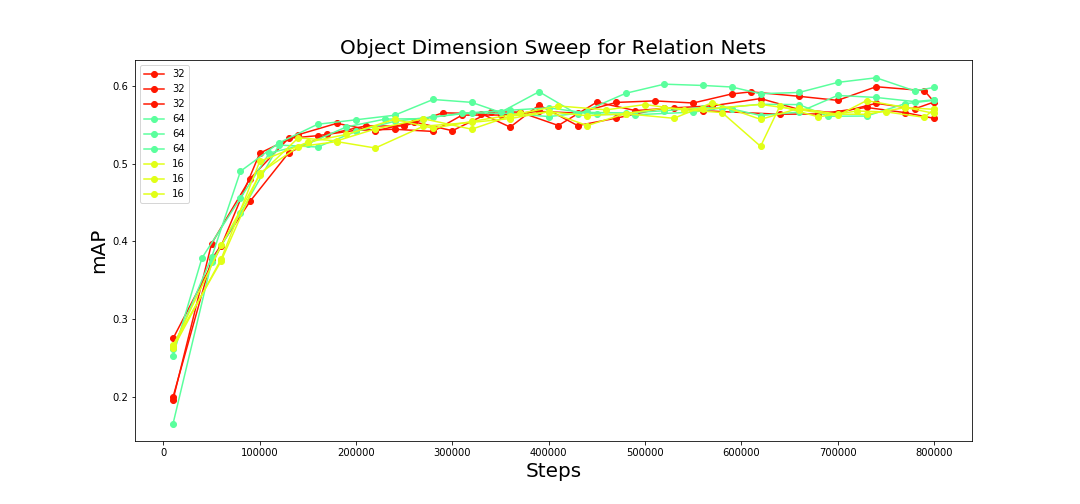}
    \caption{\map on validation for hard negatives (y-axis) vs number of training steps (x-axis) for relation network
    models on images with different dimensionality for the object embedding $o_i$.\label{fig:object_fdim_sweep}}
\end{figure}

We next show the hyperparameter sweeps for image models in deterimining the choice
of the dimensionality to represent each object $o_i$ for our image models (\cref{fig:object_fdim_sweep}).
The same choice of dimensionality was replicated for sound models.
In our initial sweeps,
on the \comp split, across
different choices of the dimensionality of objects, we found relation networks
to outperform concat and global average pooling models substantially, and thus
we picked the object dimensions based on what performs best for relation networks
since overall we are interested in the best possible choice of models for a given
split and modality. Based on the results in \cref{fig:object_fdim_sweep} we picked
$o_i \in \mathbb{R}^{64}$.

\begin{figure}
    \centering
    \includegraphics[width=\textwidth]{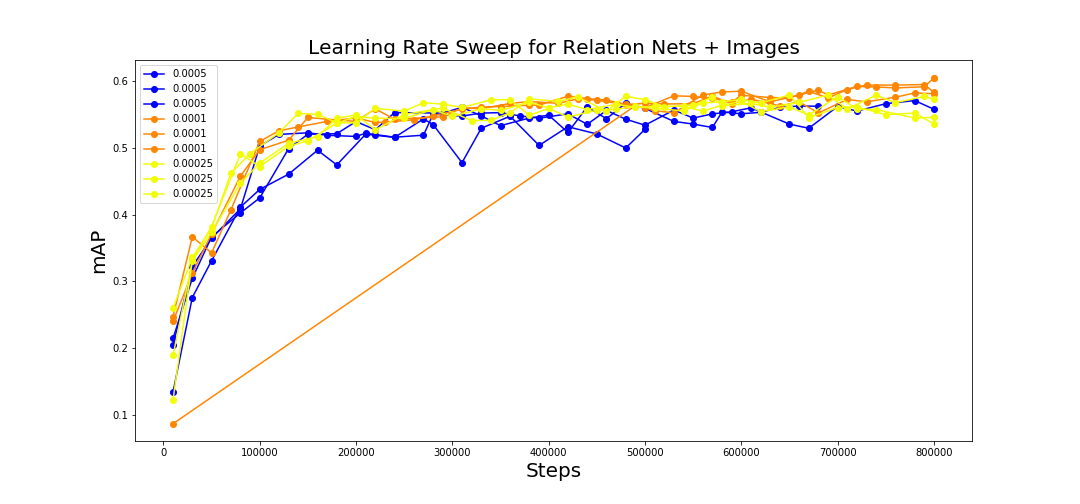}
    \caption{\map on validation for hard negatives (y-axis) vs number of training steps (x-axis) for relation network
    models on images with different learning rates.\label{fig:lr_sweep_relnets_images}}
\end{figure}

\subsection{Image relation networks learning rate sweeps}
We picked the learning rate for image models based on the best performing image relation network
model, which across an initial sweep we found to yeild the best class of models. \cref{fig:lr_sweep_relnets_images}
shows the performance of the models across learning rates of \texttt{\{1e-4, 5e-4, 2.5e-4\}}.

\subsection{Sweep on use of Language}\label{subsec:languageuse}
\begin{figure}
    \centering
    \includegraphics[width=\textwidth]{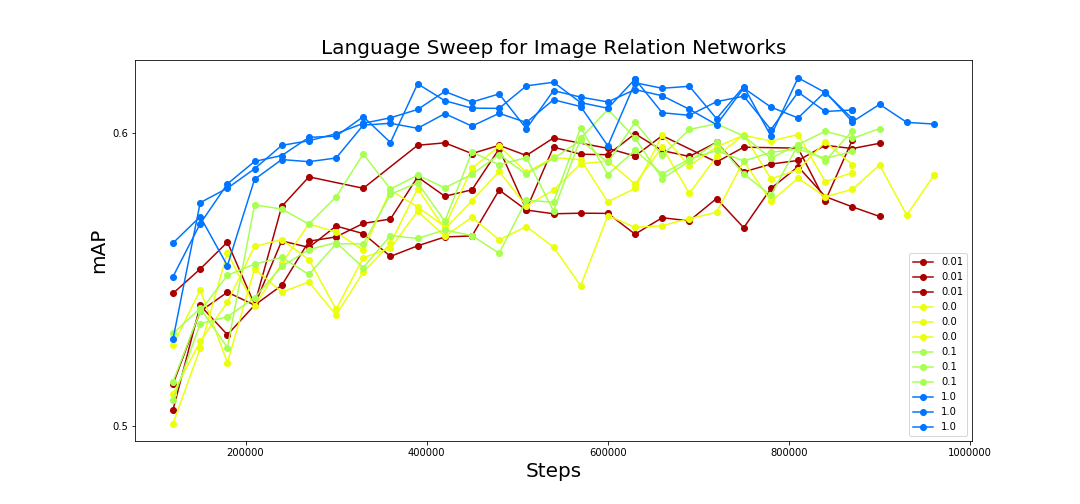}
    \caption{\map on validation for hard negatives (y-axis) vs number of training steps (x-axis) for relation network
    models on images with different amounts of language usage by varying the parameter $\alpha$.\label{fig:alpha_sweep_relnets_images}}
\end{figure}

\begin{figure}
    \centering
    \includegraphics[width=\textwidth]{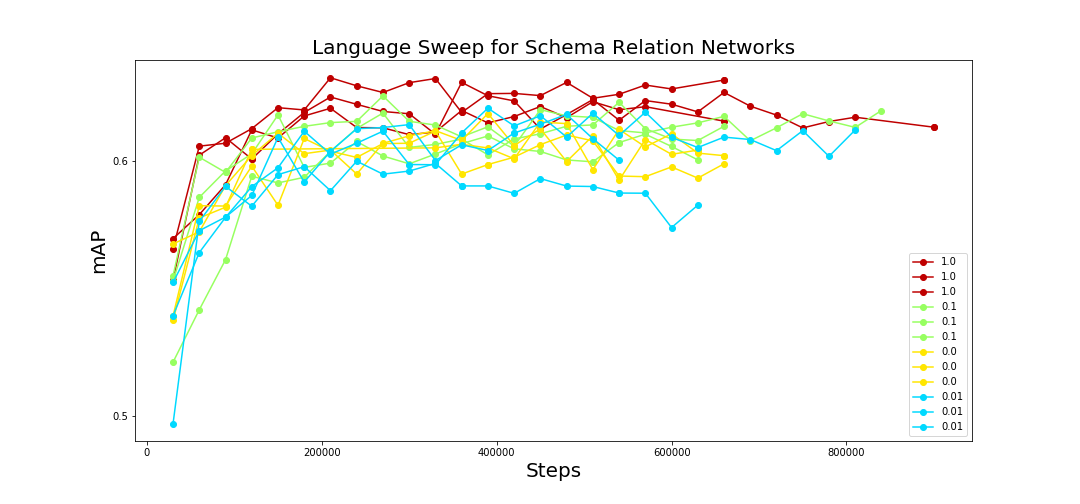}
    \caption{\map on validation for hard negatives (y-axis) vs number of training steps (x-axis) for relation network
    models on schemas with different amounts of language usage by varying the parameter $\alpha$.\label{fig:alpha_sweep_relnets_schema}}
\end{figure}

\begin{figure}
    \centering
    \includegraphics[width=\textwidth]{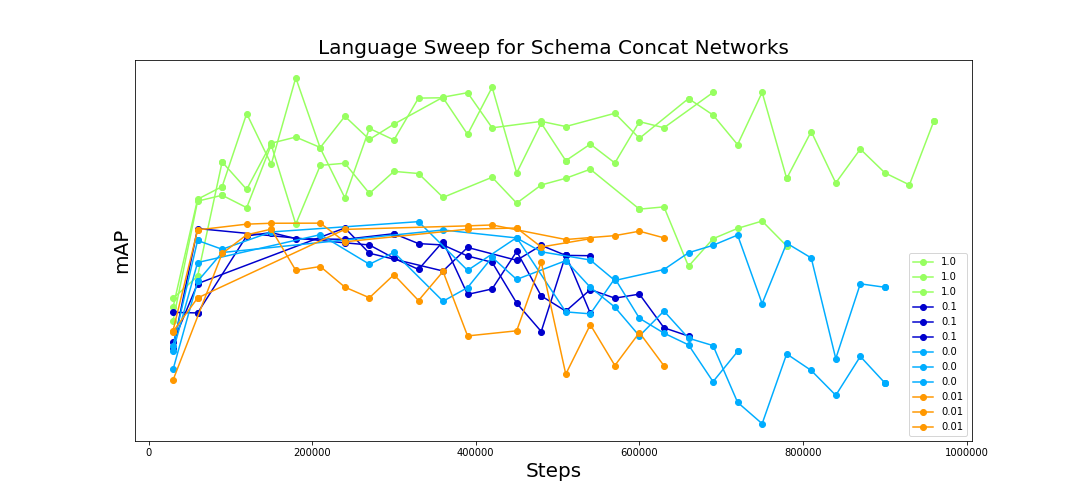}
    \caption{\map on validation for hard negatives (y-axis) vs number of training steps (x-axis) for concat pooling
    models on schemas with different amounts of language usage by varying the parameter $\alpha$.\label{fig:alpha_sweep_concat_schema}}
\end{figure}

As explained in the main paper (Figure. 4), the parameter $\alpha$ controls the tradeoff between
the query accuracy and the likelihood of the concept expressed as a prefix string. We generally
found across a broad range of values in \texttt{\{0.0, 0.01, 0.10, 1.0\}} the models generally
performed the best at $\alpha=1.0$. Our initial experiments with $\alpha=10.0$ suggested substantially
worse performance so we discareded it from the sweep. See \cref{fig:alpha_sweep_concat_schema,fig:alpha_sweep_relnets_images,fig:alpha_sweep_relnets_schema}
for the corresponding results.

\subsection{Results on Easy Negatives}\label{subsec:easynegatives}
\begin{figure}
    \centering
    \includegraphics[width=\textwidth]{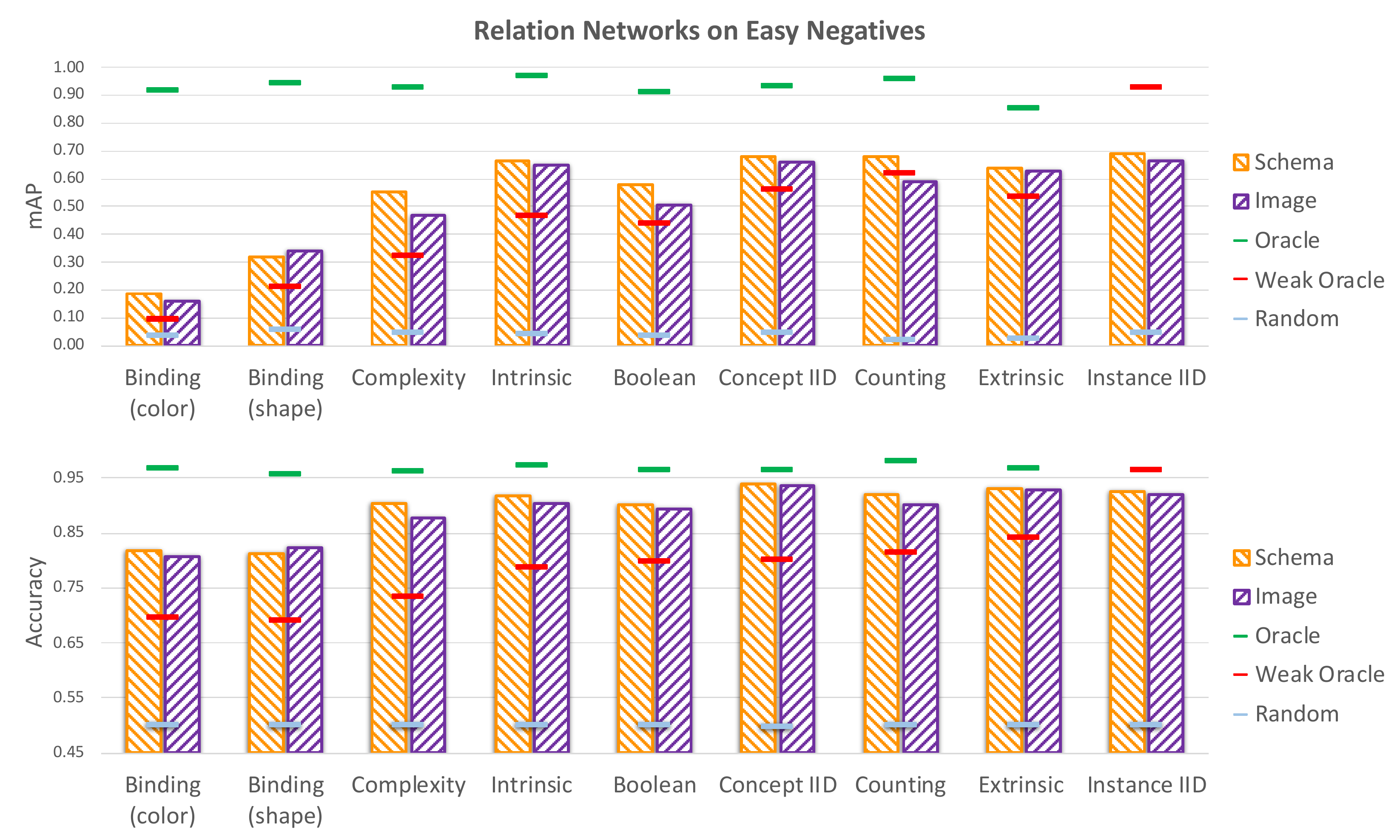}
    \caption{\map (top) and accuracy (bottom) metrics for the different splits presented in the paper
    when easy negatives are used (ordered by their corresponding \cgap). \textbf{Yellow:} shows the
    schema relation-net model while \textbf{purple:} shows the image relation-net model.
    Notice that compared to Fig. 5 in the main paper, the \cgap is smaller and the models appear
    to be substantially closer to the strong oracle in this setting compared to when we use
    hard negatives.\label{fig:rel_net_random}}
\end{figure}

In \cref{fig:rel_net_random} we show results for the \relnet model on various splits, where
easy negatives are used to populate the support and query sets during training and evaluation,
unlike the case of hard negatives discussed in the main paper (Figure. 5). Notice that the compositionality
gap (\cgap) is lower in general for easy negatives compared to the hard negatives as reported in the
main paper. Further, we find that the best models are substantially closer to the strong oracle 
compared to Figure. 5 main paper, showing that on the easier, less compositional task it is 
easier for machine learning models to approach the strong oracle (especially in terms of accuracy).
Finally, it is interesting to note that with
easy negatives it appears that the best models
outperform the weak oracle on the \counting split,
while with the hard negatives one finds that the models
are worse than the weak oracle, suggesting poor
generalization for counting.

\subsection{Finer $\alpha$ sweep for \counting}
Finally, we ran a finer alpha sweep for the \counting split since it appeared on our initial sweep
that the counting split was not performing better with language. Concretely, we ran a new set of
experiments sweeping over $\alpha$ values of \texttt{\{0.01, 0.10, 1.0, 5.0, 10.0, 100.0\}}. Across
this broader range of values, we found models still did not show any statistically significant
gains from using language \versus not for the \counting split.

\subsection{Choice of metric: \map \versus accuracy}
In general, the \map metric
opens up a larger \cgap for the various splits than indicated by
\cba. For example, with hard negatives,
while \cba indicates a gap of 14.2\% for \counting
compared to 0\% for \iid, \map suggests a gap of 34.4\% for \counting
relative to 0\% for \iid. For the \bindcolor split its 86.5\% \cgap
(\map) \versus 34.0\% for \cba. \map, while being more expensive to compute
evaluates more thoroughly
to test if a concept $h$ is truly learnt by the model, by probing
its performance on a large, representative
set of negatives $\calT$, providing
a more stringent test of compositional generalization.

\subsection{Detailed Results on all the splits in hard negatives setting}\label{subsec:full_results}
\rviclr{In this section we provide the full results of all of the tested models
on each of the splits considered in the paper, in the hard negatives setting.
Tables 1-9 show the results of different models (sorted in a descending order
based on \map for each of the splits considered in the paper, in the case where
models do not have access to language.}

\begin{table}
\centering
\caption{Performance on meta-test, sorted based on \map (in \%) on \bindcolor with hard negatives}
\begin{tabular}{llrr}
\toprule
modality &  pooling &  \map (mean) &  \map (std) \\
\midrule
    schema &    transformer &      15.9 &           0.9 \\
    schema &  relation-net &      15.3 &           0.9 \\
   image &      avg-pool &      15.1 &           0.4 \\
   image &  relation-net &      14.8 &           0.7 \\
    schema &      avg-pool &      14.4 &           1.2 \\
   image &    transformer &      14.2 &           0.2 \\
    schema &   concat &      14.0 &           0.7 \\
   image &   concat &      13.5 &           0.4 \\
   sound &      avg-pool &       9.4 &           0.6 \\
   sound &  relation-net &       9.2 &           0.3 \\
   sound &   concat &       8.0 &           0.6 \\
\bottomrule
\end{tabular}
\end{table}

\begin{table}
\centering
\caption{Performance on meta-test, sorted based on \map (in \%) on \bool with hard negatives}
\begin{tabular}{llrr}
\toprule
modality &  pooling &  \map (mean) &  \map (std) \\
\midrule
    schema &  relation-net &      51.1 &           1.4 \\
    schema &      avg-pool &      48.0 &           2.1 \\
    schema &   concat &      47.1 &           2.1 \\
    schema &    transformer &      46.9 &           1.1 \\
   image &    transformer &      44.0 &           2.2 \\
   image &  relation-net &      38.4 &           1.0 \\
   image &      avg-pool &      36.3 &           2.3 \\
   image &   concat &      29.6 &           2.4 \\
   sound &  relation-net &      25.5 &           0.8 \\
   sound &      avg-pool &      24.1 &           1.6 \\
   sound &   concat &      22.3 &           1.1 \\
\bottomrule
\end{tabular}
\end{table}

\begin{table}
\centering
\caption{Performance on meta-test, sorted based on \map (in \%) on \intrinsic with hard negatives}
\begin{tabular}{llrr}
\toprule
modality &  pooling &  \map (mean) &  \map (std) \\
\midrule
    schema &    transformer &      57.9 &           0.6 \\
    schema &  relation-net &      55.1 &           0.9 \\
   image &    transformer &      54.1 &           3.0 \\
    schema &      avg-pool &      53.3 &           3.0 \\
    schema &   concat &      52.2 &           3.5 \\
   image &  relation-net &      47.6 &           7.5 \\
   image &      avg-pool &      34.5 &           0.4 \\
   image &   concat &      34.0 &           1.0 \\
   sound &   concat &      29.4 &           1.6 \\
   sound &      avg-pool &      26.5 &           2.3 \\
   sound &  relation-net &      24.3 &           2.4 \\
\bottomrule
\end{tabular}
\end{table}

\begin{table}
\centering
\caption{Performance on meta-test, sorted based on \map (in \%) on \comp with hard negatives}
\begin{tabular}{llrr}
\toprule
modality &  pooling &  \map (mean) &  \map (std) \\
\midrule
   image &    transformer &      60.8 &           0.4 \\
    schema &  relation-net &      60.7 &           0.3 \\
    schema &    transformer &      59.8 &           0.2 \\
   image &  relation-net &      58.5 &           0.7 \\
   image &      avg-pool &      56.7 &           1.0 \\
    schema &      avg-pool &      54.6 &           1.3 \\
    schema &   concat &      53.3 &           0.9 \\
   image &   concat &      53.0 &           0.6 \\
   sound &   concat &      23.6 &           2.7 \\
   sound &      avg-pool &      22.7 &           2.3 \\
   sound &  relation-net &      21.8 &           NaN \\
\bottomrule
\end{tabular}
\end{table}

\begin{table}
\centering
\caption{Performance on meta-test, sorted based on \map (in \%) on \iid with hard negatives}
\begin{tabular}{llrr}
\toprule
modality &  pooling &  \map (mean) &  \map (std) \\
\midrule
    schema &  relation-net &      63.9 &           0.4 \\
   image &    transformer &      61.5 &           0.9 \\
    schema &    transformer &      59.8 &           0.5 \\
   image &  relation-net &      58.6 &           1.3 \\
   image &      avg-pool &      58.1 &           0.7 \\
    schema &      avg-pool &      57.4 &           0.9 \\
   image &   concat &      57.1 &           0.9 \\
    schema &   concat &      57.1 &           1.4 \\
   sound &      avg-pool &      23.5 &           1.4 \\
   sound &  relation-net &      22.3 &           2.3 \\
   sound &   concat &      21.7 &           0.5 \\
\bottomrule
\end{tabular}
\end{table}

\begin{table}
\centering
\caption{Performance on meta-test, sorted based on \map (in \%) on \extrinsic with hard negatives}
\begin{tabular}{llrr}
\toprule
modality &  pooling &  \map (mean) &  \map (std) \\
\midrule
   image &    transformer &      62.1 &           0.8 \\
    schema &    transformer &      60.9 &           0.6 \\
   image &  relation-net &      60.8 &           1.1 \\
   image &   concat &      60.8 &           0.7 \\
    schema &  relation-net &      59.8 &           0.4 \\
   image &      avg-pool &      59.4 &           1.3 \\
    schema &   concat &      54.3 &           2.3 \\
    schema &      avg-pool &      53.4 &           1.5 \\
   sound &      avg-pool &      25.7 &           NaN \\
   sound &   concat &      23.2 &           1.4 \\
   sound &  relation-net &      18.3 &           1.7 \\
\bottomrule
\end{tabular}
\end{table}

\begin{table}
\centering
\caption{Performance on meta-test, sorted based on \map (in \%) on \complexity with hard negatives}
\begin{tabular}{llrr}
\toprule
modality &  pooling &  \map (mean) &  \map (std) \\
\midrule
    schema &  relation-net &      52.8 &           0.7 \\
    schema &    transformer &      49.7 &           0.6 \\
    schema &      avg-pool &      48.4 &           1.0 \\
    schema &   concat &      48.0 &           1.2 \\
   image &      avg-pool &      45.8 &           0.4 \\
   image &    transformer &      45.6 &           0.2 \\
   image &  relation-net &      45.4 &           0.4 \\
   image &   concat &      42.6 &           3.4 \\
   sound &   concat &      26.0 &           NaN \\
   sound &  relation-net &      23.3 &           1.0 \\
   sound &      avg-pool &      23.0 &           1.7 \\
\bottomrule
\end{tabular}
\end{table}

\begin{table}
\centering
\caption{Performance on meta-test, sorted based on \map (in \%) on \bindshape with hard negatives}
\begin{tabular}{llrr}
\toprule
modality &  pooling &  \map (mean) &  \map (std) \\
\midrule
    schema &    transformer &      30.9 &           0.6 \\
    schema &  relation-net &      30.4 &           1.5 \\
   image &  relation-net &      29.1 &           1.8 \\
   image &      avg-pool &      28.7 &           1.0 \\
   image &    transformer &      28.7 &           0.1 \\
    schema &      avg-pool &      28.3 &           1.9 \\
   image &   concat &      27.0 &           0.8 \\
    schema &   concat &      27.0 &           0.8 \\
   sound &  relation-net &      17.5 &           0.4 \\
   sound &   concat &      17.4 &           1.3 \\
   sound &      avg-pool &      16.2 &           1.6 \\
\bottomrule
\end{tabular}
\end{table}

\begin{table}
\centering
\caption{Performance on meta-test, sorted based on \map (in \%) on \counting with hard negatives}
\begin{tabular}{llrr}
\toprule
modality &  pooling &  \map (mean) &  \map (std) \\
\midrule
    schema &  relation-net &      57.4 &           4.8 \\
    schema &      avg-pool &      55.0 &           6.4 \\
    schema &   concat &      50.0 &           3.2 \\
   image &      avg-pool &      48.5 &           1.3 \\
    schema &    transformer &      40.7 &           0.4 \\
   image &  relation-net &      40.3 &           2.0 \\
   image &   concat &      38.5 &           2.8 \\
   image &    transformer &      32.4 &           1.4 \\
   sound &   concat &      13.8 &           2.4 \\
   sound &      avg-pool &      13.4 &           1.0 \\
   sound &  relation-net &      13.0 &           1.7 \\
\bottomrule
\end{tabular}
\end{table}

\subsection{Detailed results on all the splits in easy negatives setting.}
\rviclr{In this section we provide the full results of all of the tested models
on each of the splits considered in the paper, in the easy negatives setting.
Tables 10-18 show the results of different models (sorted in a descending order
based on \map for each of the splits considered in the paper, in the case where
models do not have access to language. Note that we did not evaluate
\trnsf models or sound models
in this setting as this is qualitatively less interesting
than the hard negatives setting and is not the main focus of the paper.}

\begin{table}
\centering
\caption{Performance on meta-test, sorted based on \map (in \%) on \bindcolor with easy negatives}
\begin{tabular}{llrr}
\toprule
modality &  pooling &  \map (mean) &  \map (std) \\
\midrule
    schema &  relation-net &      18.6 &           1.6 \\
   image &  relation-net &      16.3 &           0.8 \\
   image &      avg-pool &      16.3 &           0.5 \\
   image &   concat &      15.6 &           0.5 \\
    schema &      avg-pool &      15.5 &           0.5 \\
    schema &   concat &      15.5 &           0.5 \\
\bottomrule
\end{tabular}

\end{table}

\begin{table}
\centering
\caption{Performance on meta-test, sorted based on \map (in \%) on \bool with easy negatives}
\begin{tabular}{llrr}
\toprule
modality &  pooling &  \map (mean) &  \map (std) \\
\midrule
    schema &  relation-net &      58.1 &           2.0 \\
    schema &      avg-pool &      55.0 &           2.1 \\
    schema &   concat &      54.7 &           2.2 \\
   image &  relation-net &      50.7 &           7.3 \\
   image &      avg-pool &      42.9 &           1.6 \\
   image &   concat &      40.9 &           1.7 \\
\bottomrule
\end{tabular}

\end{table}

\begin{table}
\centering
\caption{Performance on meta-test, sorted based on \map (in \%) on \counting with easy negatives}
\begin{tabular}{llrr}
\toprule
modality &  pooling &  \map (mean) &  \map (std) \\
\midrule
    schema &  relation-net &      67.9 &           8.5 \\
    schema &      avg-pool &      64.4 &           7.3 \\
   image &      avg-pool &      62.2 &           6.6 \\
   image &   concat &      61.2 &           6.9 \\
    schema &   concat &      60.8 &           6.7 \\
   image &  relation-net &      58.8 &           5.7 \\
\bottomrule
\end{tabular}

\end{table}

\begin{table}
\centering
\caption{Performance on meta-test, sorted based on \map (in \%) on \extrinsic with easy negatives}
\begin{tabular}{llrr}
\toprule
modality &  pooling &  \map (mean) &  \map (std) \\
\midrule
    schema &  relation-net &      63.7 &           1.4 \\
   image &  relation-net &      62.9 &           1.1 \\
   image &   concat &      61.8 &           2.4 \\
   image &      avg-pool &      61.5 &           1.4 \\
    schema &      avg-pool &      57.5 &           1.2 \\
    schema &   concat &      57.2 &           1.5 \\
\bottomrule
\end{tabular}

\end{table}

\begin{table}
\centering
\caption{Performance on meta-test, sorted based on \map (in \%) on \intrinsic with easy negatives}
\begin{tabular}{llrr}
\toprule
modality &  pooling &  \map (mean) &  \map (std) \\
\midrule
    schema &  relation-net &      66.5 &           4.8 \\
   image &  relation-net &      64.9 &           5.3 \\
   image &      avg-pool &      64.4 &           4.5 \\
    schema &      avg-pool &      63.1 &           4.7 \\
   image &   concat &      62.7 &           4.5 \\
    schema &   concat &      60.8 &           4.6 \\
\bottomrule
\end{tabular}

\end{table}

\begin{table}
\centering
\caption{Performance on meta-test, sorted based on \map (in \%) on \comp with easy negatives}
\begin{tabular}{llrr}
\toprule
modality &  pooling &  \map (mean) &  \map (std) \\
\midrule
    schema &  relation-net &      68.2 &           1.0 \\
   image &  relation-net &      65.8 &           0.4 \\
   image &      avg-pool &      65.1 &           0.4 \\
   image &   concat &      64.7 &           0.4 \\
    schema &      avg-pool &      63.3 &           0.4 \\
    schema &   concat &      61.9 &           0.2 \\
\bottomrule
\end{tabular}

\end{table}

\begin{table}
\centering
\caption{Performance on meta-test, sorted based on \map (in \%) on \iid with easy negatives}
\begin{tabular}{llrr}
\toprule
modality &  pooling &  \map (mean) &  \map (std) \\
\midrule
    schema &  relation-net &      69.0 &           3.6 \\
   image &      avg-pool &      66.8 &           3.3 \\
   image &  relation-net &      66.5 &           3.2 \\
   image &   concat &      65.4 &           3.4 \\
    schema &      avg-pool &      63.5 &           3.1 \\
    schema &   concat &      63.3 &           3.1 \\
\bottomrule
\end{tabular}

\end{table}

\begin{table}
\centering
\caption{Performance on meta-test, sorted based on \map (in \%) on \complexity with easy negatives}
\begin{tabular}{llrr}
\toprule
modality &  pooling &  \map (mean) &  \map (std) \\
\midrule
    schema &  relation-net &      55.1 &           1.8 \\
    schema &      avg-pool &      51.9 &           2.0 \\
    schema &   concat &      51.3 &           1.9 \\
   image &      avg-pool &      49.0 &           1.6 \\
   image &   concat &      48.2 &           1.7 \\
   image &  relation-net &      46.6 &           1.5 \\
\bottomrule
\end{tabular}

\end{table}

\begin{table}
\centering
\caption{Performance on meta-test, sorted based on \map (in \%) on \bindshape with easy negatives}
\begin{tabular}{llrr}
\toprule
modality &  pooling &  \map (mean) &  \map (std) \\
\midrule
   image &  relation-net &      33.9 &           2.4 \\
    schema &      avg-pool &      32.1 &           1.7 \\
   image &      avg-pool &      31.9 &           1.8 \\
    schema &  relation-net &      31.7 &           1.6 \\
   image &   concat &      31.3 &           1.7 \\
    schema &   concat &      31.0 &           1.5 \\
\bottomrule
\end{tabular}

\end{table}

%% file: sections/appendix_sections/qualitative_examples.tex
\begin{figure}[htbp]
    \includegraphics[width=\textwidth]{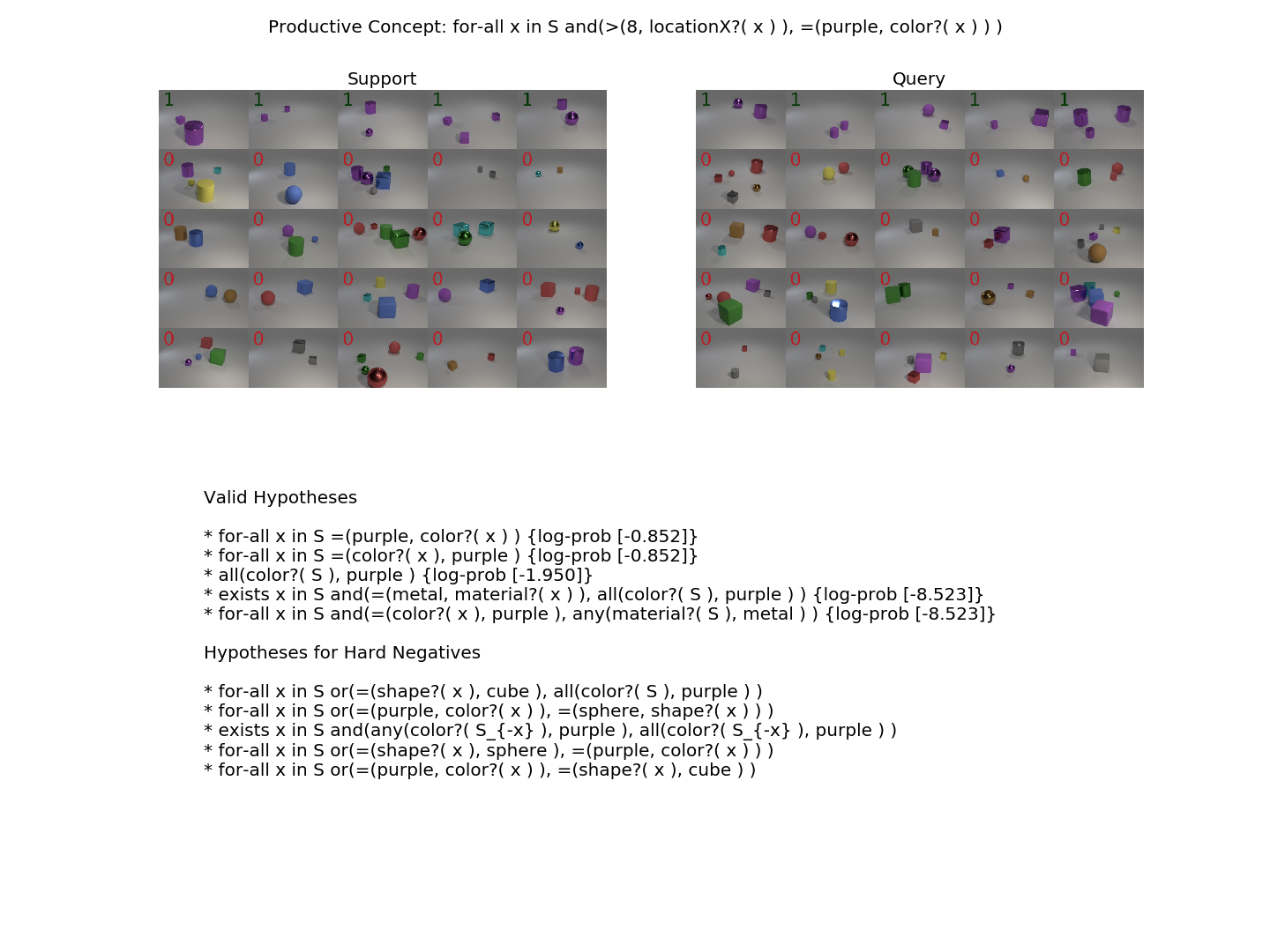}
    \caption{\label{fig:qual_1} Qualitative Example of an Episode in \pcl dataset. Best viewed zooming in, in color.}
\end{figure}

\begin{figure}[htbp]
    \includegraphics[width=\textwidth]{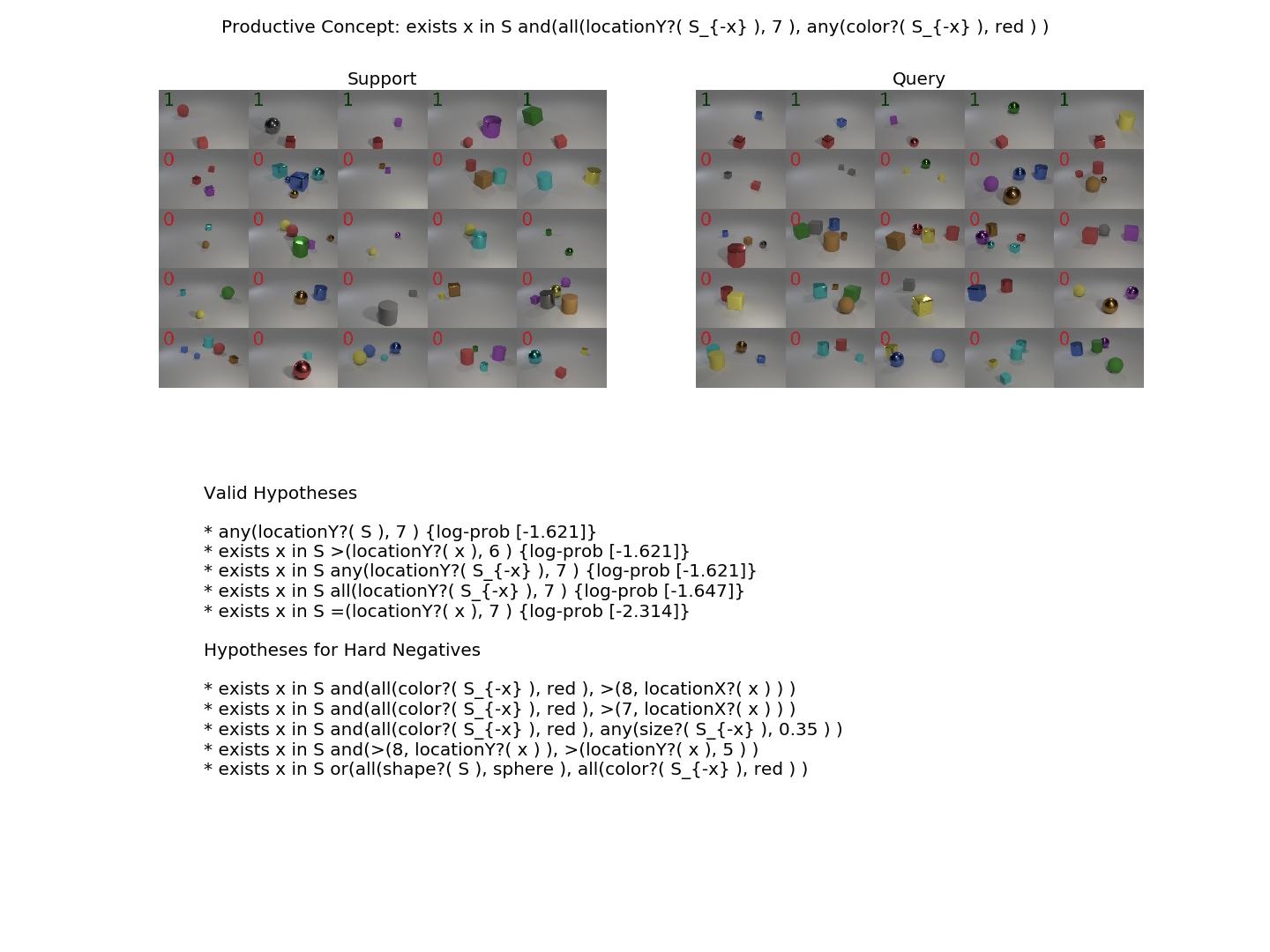}
    \caption{\label{fig:qual_2}Qualitative Example of an Episode in \pcl dataset. Best viewed zooming in, in color.}
\end{figure}

\begin{figure}[htbp]
    \includegraphics[width=\textwidth]{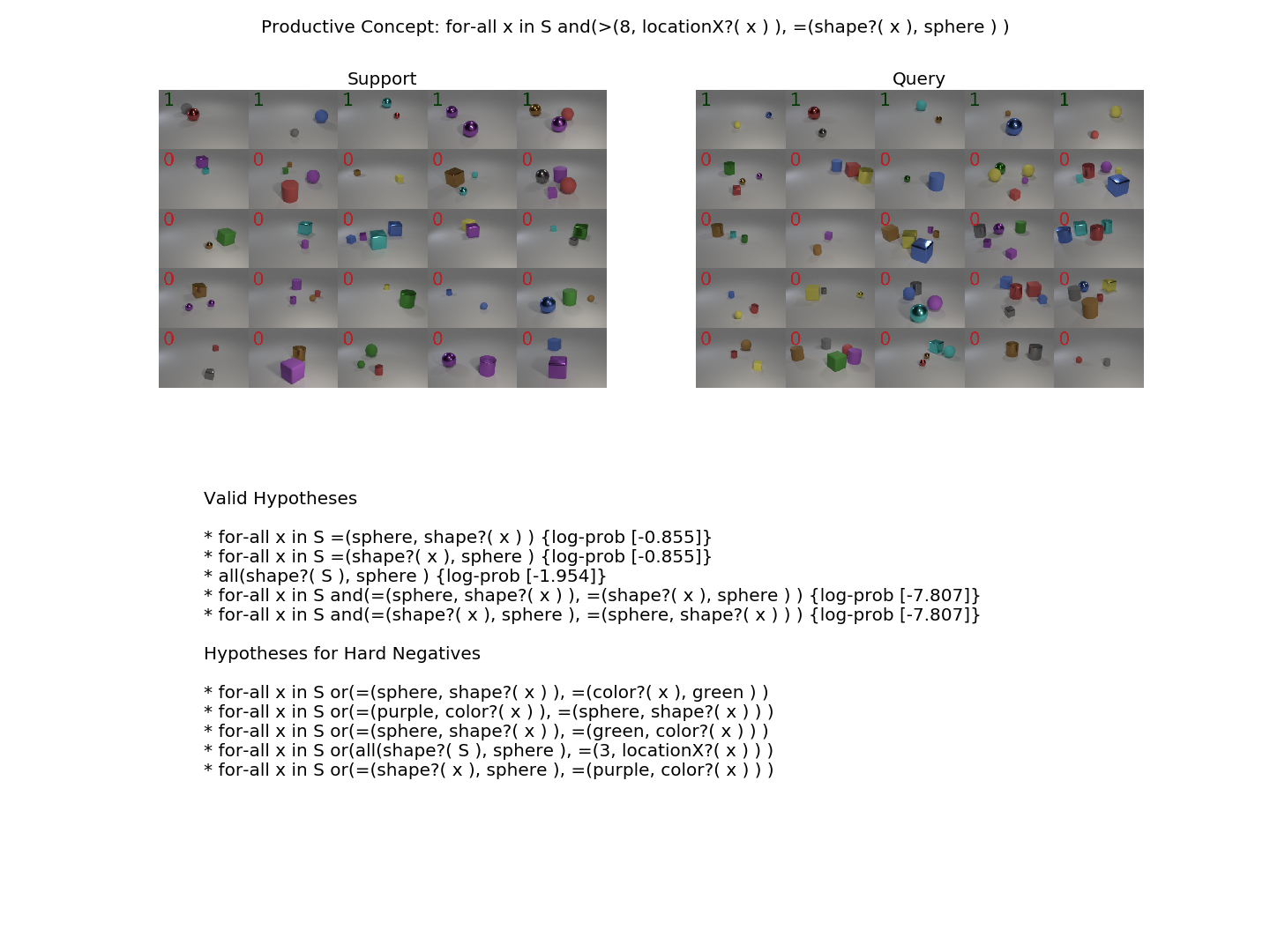}
    \caption{\label{fig:qual_3}Qualitative Example of an Episode in \pcl dataset. Best viewed zooming in, in color.}
\end{figure}

\begin{figure}[htbp]
    \includegraphics[width=\textwidth]{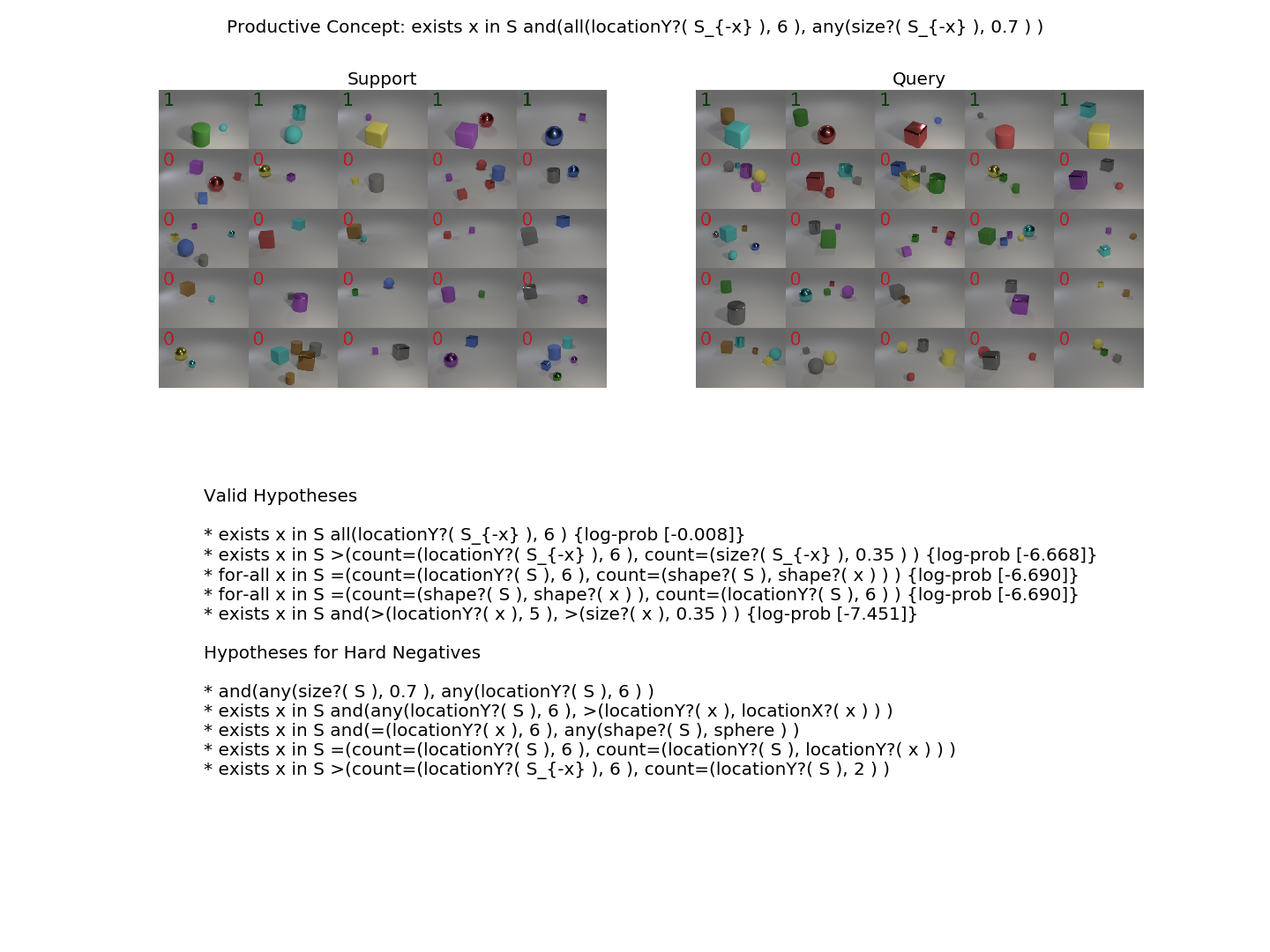}
    \caption{\label{fig:qual_4}Qualitative Example of an Episode in \pcl dataset. Best viewed zooming in, in color.}
\end{figure}

\begin{figure}[htbp]
    \includegraphics[width=\textwidth]{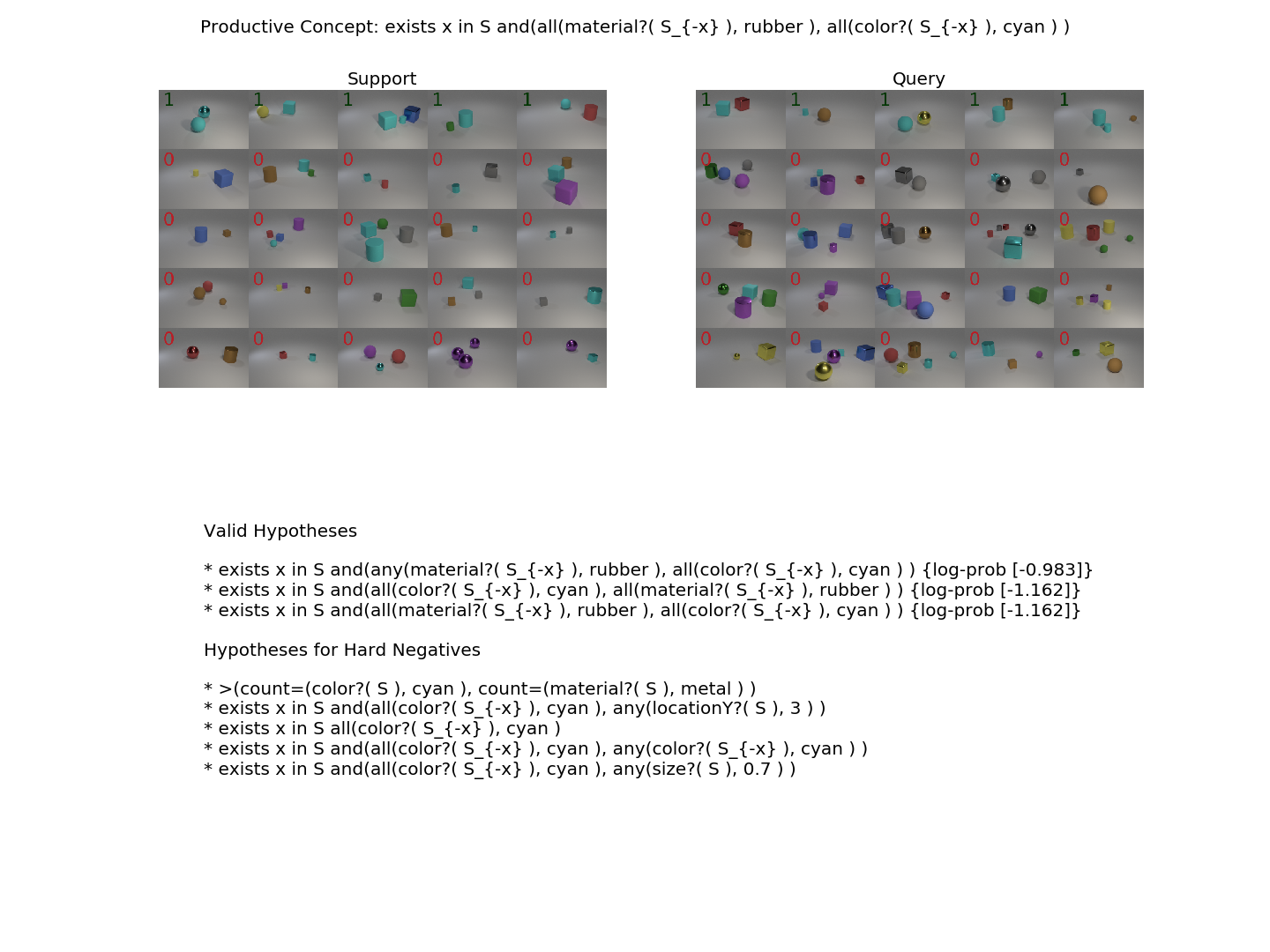}
    \caption{\label{fig:qual_5}Qualitative Example of an Episode in \pcl dataset. Best viewed zooming in, in color.}
\end{figure}

\begin{figure}[htbp]
    \includegraphics[width=\textwidth]{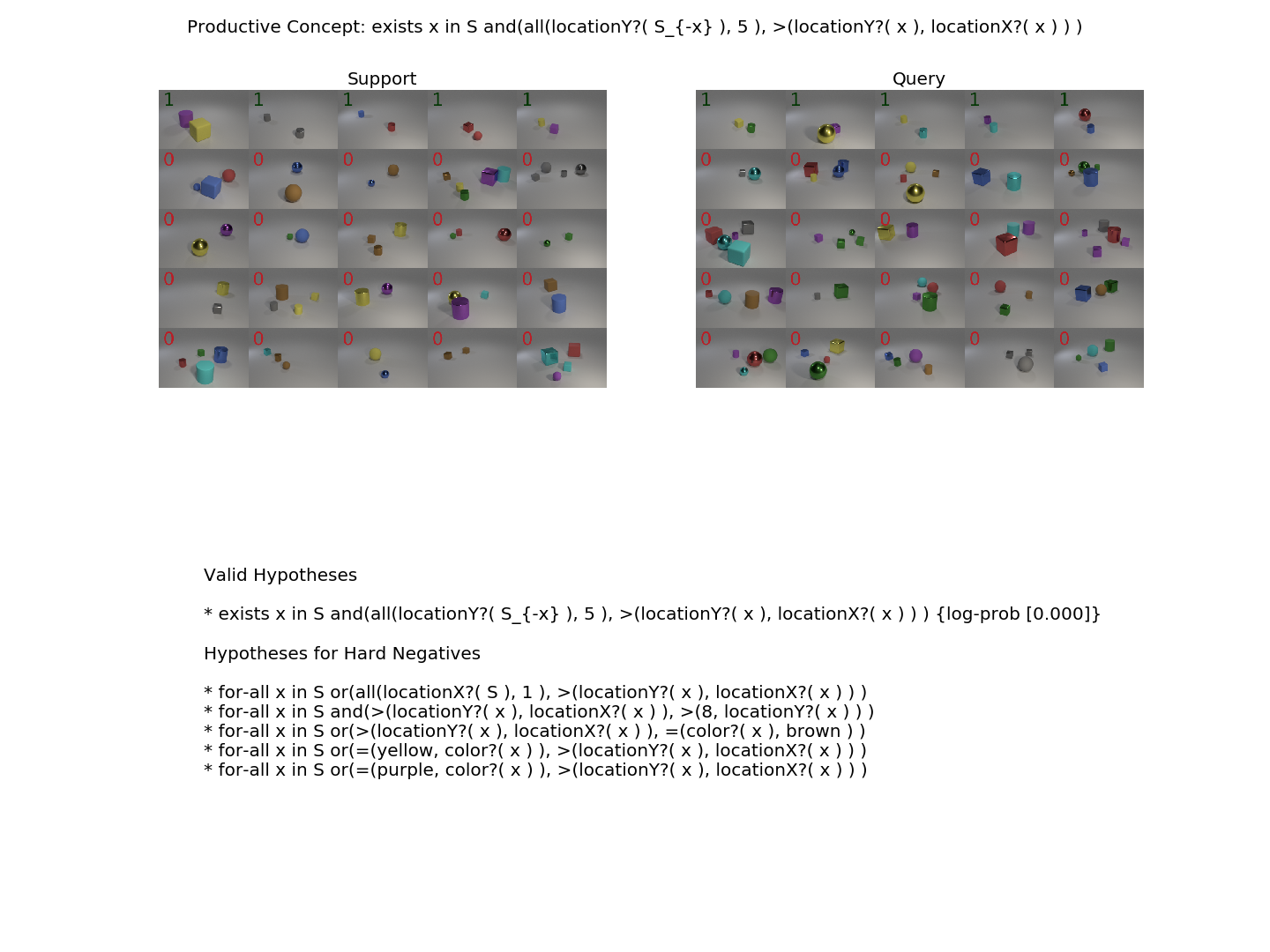}
    \caption{\label{fig:qual_6}Qualitative Example of an Episode in \pcl dataset. Best viewed zooming in, in color.}
\end{figure}

%% file: sections/appendix_sections/full_grammar.tex
\begin{lstlisting}[escapeinside={(*}{*)}]
START -> (*$\lambda$*) S. BOOL exists= | (*$\lambda$*) S. BOOL for-all=

BOOL  -> BOOL BOOL and | BOOL BOOL or | BOOL not |
  C C = | SH SH = | M M = | SI SI = | L L = |
  NUM NUM = | SI SI > | L L > | NUM NUM > |
  SETFC C all | SETFSH SH all | SETFM M all |
  SETFSI SI all | SETFL L all | SETFC C any |
  SETFSH SH any | SETFM M any | SETFSI SI any |
  SETFL L any

NUM   -> SETFC C count= | SETFSH SH count= |
  SETFM M count= | SETFSI SI count= | SETFL L count=
NUM   -> 1 | 2 | 3 

SETFC -> SET FC
SETFSH-> SET FSH
SETFM -> SET FM
SETFSI-> SET FSI
SETFL -> SET FL

C     -> gray | red | blue | green | brown | purple |
  cyan | yellow
C     -> OBJECT FC

SH    -> cube | sphere | cylinder
SH    -> OBJECT FSH

M     -> rubber | metal
M     -> OBJECT FM

SI    -> large | small
SI    -> OBJECT FSI

L     -> 1 | 2 | 3 | 4 | 5 | 6 | 7 | 8
L     -> OBJECT FL

FC    -> color?
FSH   -> shape?
FM    -> material?
FSI   -> size?
FL    -> locationX? | locationY?

OBJECT-> (*$\vx$*)
SET: (*$S$*) | (*$S_{-\vx}$*)
\end{lstlisting}

%% file: sections/appendix_sections/removed_concepts.tex
\begin{verbatim}
    exists=x \in S or(
      =(locationX?( x ), locationY?( x ) ),
      any(color?( S ), brown ) 
    )
    exists=x \in S and(
      exists=(locationY?( S ), locationX?( x ) ),
      any(color?( S ), brown ) 
    )
    exists=x \in S or(
      all(color?( S ), gray ),
      all(color?( S ), brown )
    )
\end{verbatim}

%% file: sections/appendix_sections/high_prob_prior.tex
\begin{lstlisting}
    exists x in S =(2, count=(color?( $S_{-x}$ ), cyan ) ),
    exists x in S >(locationY?( x ), 6 ),
    =(count=(color?( S ), brown ), 3 ),
    >(count=(locationX?( S ), 3 ), 2 ),
    any(locationY?( S ), 6 ),
    =(1, count=(locationY?( S ), 7 ) ),
    =(3, count=(locationY?( S ), 3 ) ),
    all(locationX?( S ), 2 ),
    exists x in S all(locationY?( $S_{-x}$ ), 5 ),
    =(2, count=(color?( S ), blue ) ),
    for-all x in S not( >(6, locationX?( x ) ) ),
    =(count=(color?( S ), gray ), 2 ),
    =(2, count=(color?( S ), gray ) )
\end{lstlisting}